\documentclass[final,1p,times]{elsarticle}

\usepackage[cm]{fullpage}
\usepackage{pdflscape} 
\usepackage{lmodern,amsmath,amssymb}
\usepackage{commath}
\usepackage{a4wide}
\usepackage{algorithm}
\usepackage{algpseudocode}
\usepackage{multirow}
\usepackage{graphicx}
\usepackage{multicol}
\usepackage{tabularx}
\usepackage{vwcol} 
\usepackage{color,soul}
\usepackage{subcaption}
\usepackage{url}
\usepackage{booktabs}
\usepackage{multirow}
\usepackage{mathtools}
\usepackage{placeins,cleveref}
\usepackage[export]{adjustbox}
\usepackage{placeins}
\usepackage{ragged2e}
\usepackage[toc,page]{appendix}
\usepackage{lipsum}
\usepackage{caption}
\usepackage{textcomp}

\biboptions{square,sort}

\journal{arXiv}

\begin{document}
	
	\begin{frontmatter}
		
		
		
		\title{\emph{CNAK} : Cluster Number Assisted \emph{K-means} }
		
		
		\author{Jayasree Saha}
		\ead{jayasree.saha@iitkgp.ac.in}
		\author{Jayanta Mukherjee}
		\ead{jay@cse.iitkgp.ac.in}
		\address{Department of Computer Science and Engineering\\Indian Institue of Technology Kharagpur\\ West Bengal, India, 721302}
		
		\begin{abstract}

Determining the number of clusters present in a dataset is an important problem in cluster analysis. Conventional clustering techniques generally assume this parameter to be provided up front. 
In this paper, we propose a method which analyzes cluster stability for predicting the cluster number. Under the same computational framework, the technique also finds representatives of the clusters. The method is apt for handling big data, as we design the algorithm using \emph{Monte-Carlo} simulation. Also, we explore a few pertinent issues found to be of also clustering. Experiments reveal that the proposed method is capable of identifying  a single cluster. It is robust in handling high dimensional dataset and performs reasonably well over datasets having cluster imbalance. Moreover, it can indicate cluster hierarchy, if present. Overall we have observed significant improvement in speed and quality for predicting cluster numbers as well as the composition of clusters in a large dataset.
	
		\end{abstract}
		
		\begin{keyword}
			\emph{k-means} clustering, Bipartite graph, Perfect Matching, \emph{Kuhn-Munkres} Algorithm, \emph{Monte Carlo} simulation.

			
		\end{keyword}
		
	\end{frontmatter}
	
\section{Introduction}
\label{sec1}
In cluster analysis, it is required to  group a set of data points  in a multi-dimensional   space,  so that data points in the same group  are more similar  to each other than to those in other groups. These groups are called clusters. Various distance functions may be used to compute the degree of similarity or dissimilarity among these data points. Typically Euclidean distance function is widely used in clustering. The aim of this unsupervised technique is to increase homogeneity in a group and heterogeneity between groups. Several clustering methods with different characteristics have been proposed for different purposes. Some well-known methods include partition-based clustering~\citep{Lloyd57}, hierarchical clustering~\citep{Hierarchy1963}, spectral clustering~\citep{onspectral2001}, density-based clustering~\citep{DBSCAN}. However, they require the knowledge of cluster number for a given dataset a priori~\citep{Lloyd57,onspectral2001,
	DBSCAN,DBCLASD,DENCLUE}.
Nevertheless, the estimation of the number of clusters is a difficult problem as the underlying data distribution is unknown. Readers can find several existing techniques for determining cluster number in~\citep{survey_cluster_number2017,R3_Chiang2010}. We have followed the nomenclature used in \cite{R3_Chiang2010} for categorizing different methods for the prediction of cluster number.
In this work, we choose to focus only on three approaches: 1) Variance-based approach, 2) Structural approach, and 3) Resampling approach.
Variance-based approaches are based on measuring compactness within a cluster. Structural approaches include between-cluster separation as well as within cluster variance. We have chosen these approaches as they are either more suitable for handling big data, or  appear in a comparative study by several researchers. Some well known approaches are Calinski-Harabaz~\citep{CH}, Silhouette Coefficient~\citep{sil}, Davies-Bouldin~\citep{DB}, Jump~\citep{jump}, Gap statistic~\citep{gap}, etc. These approaches are not appropriate for handling the big data, as they are computationally intensive and require large storage space. It requires a scalable solution~\citep{kluster2018, ISI_LL_LML2018} for identifying the number of clusters. Resampling-based approaches can be considered in such a scenario. Recently, the concept of stability in clustering has become popular. A few methods  \citep{instability2012, CV_A} utilize the concept of clustering robustness against the randomness in the choice of  sampled datasets to explore clustering stability. 
In the following subsection, we discuss our motivation for this work.
\subsection{Motivation}
In general, clustering algorithms deal with two problems:
1) estimation of natural cluster number, and
2) determination of well-organized groups.
In the current scenario, many applications requiring the task of clustering target a large size of data. Therefore, a scalable solution is required for determining  cluster number and well-formed groups.
There are clustering methods~\citep{Davidson2003}, which employ \emph{Monte-Carlo} based techniques to perform clustering on a large dataset. But, they require prior information about the cluster number. Also, there is no such built-in method to select the size of a sampled dataset in \emph{Monte-Carlo} based methods. Existing methods usually sample a large fraction of dataset, and use them for clustering. Hence, the improvement with respect to computational time for those algorithms is limited. The objectives for this work are as follows: 
1)	estimation of cluster number, 
2)	simultaneous computation of cluster representatives during the process of estimation, and
3)	estimation of the required  size for a sampled dataset in \emph{Monte-Carlo} Simulation.
Our contributions in this work are summarized below.
\begin{itemize}
	\item We propose a {\emph Monte-Carlo} based technique to determine appropriate cluster number for a given dataset. We derive cluster representatives from the same method.
	
	\item We propose a model for estimating the size of the sampled dataset for the {\emph Monte-Carlo} based simulation. Also, we test the relevance of the solution empirically with many datasets.
	
	\item We demonstrate  the applicability of our proposed method for large-scale datasets.
	
	\item We explore the behavior of our method in a few relevant issues in clustering, such as 1) cluster overlapping, 2) shape clusters, and 3) clustering in the presence of noise.
	
\end{itemize}

\section{Proposed Work}
It comprises of two steps. First, it estimates the cluster number. Subsequently, it presents the clustering technique which provides support towards the correctness of cluster number prediction.
\subsection{Cluster Number Prediction}
We use a fraction of the original dataset to predict cluster number.  Initially,
we randomly select a sampled dataset of size $\frac{1}{s}$ of the original dataset. We prepare $T$ such random sampled sets. Then, we generate $k$ cluster centers by applying \emph{k-means++} \citep{Arthur:2007} algorithm to each sampled datasets. Hence, we have $T$ instances of cluster centers of the original dataset. We perform a bipartite matching between each pair of $k$ centroids, generated from each pair of $T$ sampled datasets. The central idea is as follows:

If natural cluster number is chosen for clustering, 
then every \emph{k-means} clustering on different sampled dataset should have near-by cluster centers. Hence, the average distance between matched centroids of each pair of the sampled dataset should be minimum for the true cluster number. 
Thus, if we plot such a distance in a normalized form against $k$ that varies from $2$ to any higher positive integer, we should get a local minimum in the plot near the true cluster number for an underlying dataset.
It is recommended to take a range of cluster numbers to compute these normalized distances if any knowledge about it exists for a given dataset.

\subsubsection{Problem Formulation}
We have designed our problem in two steps. The first step describes one to one mapping of cluster centers from sampled dataset $X$ to sampled dataset $Y$. The second step computes a score which has a direct impact in selecting true cluster number. \\
{\em Step 1} : 
Let $Y = \{y_{1}, \cdots, y_{N} \}$ be the original dataset and Let $S =\{S_{1}, \cdots, S_{T} \}$ be a set of random sampled data points. Each element $S_{i}$ is of size $\frac{1}{s}$ of the original dataset $Y$. Each sampled dataset is prepared by sampling without repetition. Let $R=\{C_{i}=\{c_{1}, \cdots, c_{k}\}\}_{i=1}^{T}$ be the set of cluster representatives of $S$. We form a complete bipartite graph by using any $C_{x}$ and $C_{y} \in R$. Let $G=(V,E)$ be a complete weighted bipartite graph with weight function $\mathcal{W}: E \rightarrow \mathbb{R}$ such that $V=(C_{x} \cup C_{y})$ and $E=\{(u,v) : u \in C_{x},  v\in C_{y}\}$. The weight function measures the \emph{Euclidean} distance between two cluster centroids. The goal is to find the minimum perfect bipartite matching $\mathcal{M}_{xy}$. We have used {\em Kuhn-Munkres}~\cite{kuhn-munkre} algorithm to find the solution for the above goal.
This method solves the problem in time $\mathcal{O}(k^{3})$. \\
{\em Step 2} :
In this step, we formulate a score ($\mathbb{S}_{j}$) for cluster number $j$. We have $T$ sampled datasets. Hence, we have $\binom{T}{2}$ number of mappings. We compute the cost ($\mathcal{C}_{xy}$) of mapping $\mathcal{M}_{xy}$ as the sum of its edge weights. However, the score for cluster number $j$ can be defined as follows:
\begin{equation}
\label{eq:score}
\begin{array}{llll} \mathbb{S}_{j}&=&\frac{2}{T(T+1)}(\sum_{x=1}^{T}\sum_{y=1,y\neq x}^{T} \mathcal{C}_{xy})
\end{array}
\end{equation}
We generate scores for $j=k_{1}$ to $k_{2}$. In our experiment, we use $k_{1}=1$ and $k_{2}=30$. The optimal $k^{*}$ is selected as follows:
\begin{equation}
\begin{array}{llll}
k^{*} &=&\underset{j}{\mathrm{argmin}} &\mathbb{S}_{j}
\end{array}
\end{equation}
\subsection{Clustering}
We collect each $\mathcal{M}_{xy}$ mappings for cluster number $k^{*}$.
We call it bucketization. In our work, we choose a sampled dataset (say, $S_{1}$) as reference. We use mapping of cluster centroids of  other sampled datasets with the reference cluster ($S_1$) to bag similar centroids. We consider a bucket with $k^{*}$ cells. We push similar centroids into one cell. The mean of each cell of the bucket is evaluated to provide $k^{*}$ representatives for the dataset. Algorithm~\ref{algo:Algo1} and Algorithm~\ref{algo:Algo2} together describe the \emph{cluster number assisted k-means (CNAK)} algorithm.

\subsection{Parameter selection}
We performed $T$ number of random trials  (empirically chosen as 50). It is expected that estimated values of k (the cluster number) would lie within a narrow interval. If it is not so, we may increase this number and iterate this step until an appropriate $T$ is found.
\begin{algorithm}[H] 
	\caption{Cluster Number Assisted k-means (CNAK)}
	\label{algo:Algo1}
	\begin{algorithmic}[1]
		\Require{ Y, $\gamma$, T}
		\Ensure{ C,  P}\\
		
		{Description: Y=dataset; $\gamma$=size of sampled dataset; T=Number of samples;}
		{C=cluster representatives; P=partition of clusters.}
		\For{$i\gets k_{1}$ to $k_{n}$} 
		
		\For{$j\gets1$ to $T$ }
		\State	S(j)=Sampling $\gamma$ points from Y.
		\State	C(j)=k-means++(data=sample(j), k=i)
		\EndFor
		
		\For{each pair of sampled datasets}
		\State	Create complete bipartite graph. 
		\State  Execute kuhn-Munkres algorithm.
		\State  Store mapping and cost for this pair.
		\EndFor
		
		\State Compute average score for cluster number i. 
		\State Select $k^{*}=i$ where $i^{th}$ average score is minimum.
		\State Call Bucketization(C, mapping, $k^{*}$, T).
		\State Store $k^{*}$ cluster centroids retured by Bucketization.
		\State Create partition by assigning data points to the nearest cluster centroid.
		\EndFor
		
	\end{algorithmic}
\end{algorithm}	

\begin{algorithm}[H] 
	\caption{Bucketization}
	\label{algo:Algo2}
	\begin{algorithmic}[1]
		\Require{centroids, mapping, $k$, $T$}
		\Ensure{Cluster Centers C}
		\Statex
		\State Initialize bucket $B$ with $k$ cells.
		\State Select $k$ cluster centroids of $S_{1}$ (reference) sampled dataset.
		\State Push $k$ cluster centroids in $k$ cells of the bucket.
		\For{$i \gets 2$ to $T$}
		\State	M=Mapping between cluster centroids of $S_{1}$ and $S_{i}$.
		\State	Push cluster centroids into bucket B using mapping M.
		\EndFor
		
		\For{$j \gets 1$ to $k$}
		\State	C(j)=Compute mean of all items in $j^{th}$ cell of $B$.
		
		\EndFor
	\end{algorithmic}
\end{algorithm}

\section{Estimation of sample size}\label{sec:sample_size}
Size of sampled dataset influences  any statistical estimate directly. In general, large sample size brings better estimates. There are many applications which require the determination of the size of a sampled dataset. e.g., factor analysis, regression analysis, cluster analysis, etc. However, no such method exists for computing size of a sampled dataset for cluster analysis. Although we can consider any large value for the size of a sampled dataset, this arbitrary selection becomes vulnerable with increasing size of the dataset for cluster analysis. Here, we adopt the {\emph central limit theorem}~\cite{SS} to estimate the size. However, it comes with the condition that the sample size should be sufficiently large.
We discuss a theory adopted in this paper for determining the size of a sampled dataset, followed by the description of a heuristic to estimate its parameters.

\subsection{Theory}
Let a sampled dataset of size $\gamma$ be drawn randomly without replacement from a population of size $N$. Let $X$ be a random variable that represents a $1$-dimensional feature vector of the population. Let the population mean be $\mu$, and the standard deviation be $\sigma$. According to the sampling theorem~\citep{sampling,SS} the expected value of mean and variance of the probability distribution of sample mean ($\bar{X} =\frac{1}{\gamma}\sum_{i=1}^{\gamma}X_{i}$) are given by:
\begin{equation}
\begin{array}{lllllll}
E(\bar{X})&=&\mu&,&\sigma_{\bar{X}}^{2}&=&\frac{\sigma^{2}}{\gamma}\frac{N-\gamma}{N-1}
\end{array}
\label{eq:mean-variance}
\end{equation}

According to \emph{central limit theorem} for large $N$ and $\gamma$, the probability distribution of $\bar{X}$ is approximately normal with mean and variance are given by Eq.\eqref{eq:mean-variance}. Therefore, the distribution of the ratio $\frac{\bar{X}-\mu}{\sigma_{\bar{X}}}$ is a normal distribution. 
Thus, the size of a sampled dataset for without replacement($\gamma$) sampling strategy~\cite{sampling} is given by:
\begin{equation}
\label{eq:sample_size}
\begin{array}{lllllll}
\gamma_{1}&=&(\frac{Z_{\frac{\beta}{2}}}{c})^{2}\sigma^{2}&,&\gamma&=&\frac{\gamma_{1}}{1+(\frac{\gamma_{1}}{N})}
\end{array}
\end{equation}
where $Z_{\frac{\beta}{2}}$ is the probability of a random
variable following a normal distribution with $0$ mean and variance $1$, within the interval $[\frac{\beta}{2} , \infty]$. $c$ is the marginal error. so that the sample mean $\bar X$ lies in the interval of $\mu-c$ to $\mu+c$ , where $\mu$ is the population mean.

\subsection{{A heuristic on determination of the size of sampling in a multdimensional space}} \label{sub:sample_size}
For the multidimensional data point, we perform eigen analysis of its covariance matrix and consider that the maximum eigen value $\lambda_{max}$ is a measure of the variance of the data and is related to $\sigma^2$. Here, $\lambda_{max}$ is chosen to capture the maximum possible variance in the dataset. Choice of values of ($Z_{\frac{\beta}{2}}, c, \lambda_{k}$) is discussed below.

We attempt to keep the size of the sampled dataset as small as possible. The parameter $c$ in Eq.~\ref{eq:sample_size} plays a key role here. It measures the amount of variance we can allow to predict the population mean. Empirically, we find a log-linear relationship between $c$ and $\lambda_{max}$ where  $\lambda_{max}$ represents the maximum eigen value. We follow Eq.\ref{eq:c} to determine the size of a sampled dataset.

\begin{equation}
\label{eq:c}
\begin{array}{lllllll}
c&=&0.6&\text{if}&\lambda_{1}<60&\text{and}&\lambda_{1}-\lambda_{2}>\epsilon\\
c&=&0.2&\text{if}&\lambda_{1}<60&\text{and}&\lambda_{1}\approx\lambda_{2}\\
c&=&(\lambda_{max})^{\frac{1}{\tau}},&\text{otherwise} &&&\text{where $\tau$ is an integer} 
\end{array}
\end{equation}
We have tabulated the size of the sampled dataset and the parameters $(\lambda_{max}, \tau, c)$ in Tables \ref{tab:SS_size}, \ref{tab:dims}, \ref{tab:shape} and Table \ref{tab:DD} for the synthetic and real-world dataset, selected for our experiments,  respectively. We set $\epsilon=10$ empirically.

\subsection{Observations}
We observe that our heuristics show reasonable estimates for large scale data compared to other small scale datasets, as shown in Table~\ref{tab:SS}. This enables the computation time of our proposed method very less compared to other methods. However, we find that the heuristic fails when many clusters are present in the dataset such as \emph{sim-9}. We consider $70\%$ of the original dataset as the size of the sampled dataset for this dataset.

\section{Comparative Study}\label{sec:compM}
Several works have been reported for estimating the number of clusters in a dataset. They are categorized broadly into five types~\cite{R3_Chiang2010}. We choose three of them for the comparative study: 1) Variance based approach, 2) Structural approach and 3) Resampling approach. We choose the first two approaches due to their wide popularity, whereas the third approach is capable of handling large scale data. Although the first two types contain a wide variety of strategies for finding cluster number, we have selected a few methods which have been cited and used in comparisons in a good number of woks. We consider them in our comparative study. However, determination of the optimal number of clusters does not reflect on the quality of clustering. Hence, a thorough analysis of the partition with the optimal cluster number is needed. Our proposed approach finds the cluster number as well as partitions in the dataset. We choose clustering techniques for comparative study based on the following criterion: 1) suitability of handling a large scale dataset (RPKM~\cite{CAPO2017}),  2) effectiveness of resampling-based methods for clustering (Bootstrap Averaging~\cite{Davidson2003}),  and 3) effective initial choice of centroids in \emph{k-means} based clustering algorithm (Refining Seed~\cite{Bradley1998RefiningIP})

\subsection{Baseline Approaches for Cluster Number Determination}\label{sub:BCN}
The Curvature~\cite{Curvature} method, chooses the number of clusters which maximizes the following measure.
\begin{equation}
\label{eq:curv}
\tau_{k}= \dfrac{J(k-1)-J(k)}{J(k)-(k+1)} \\
\end{equation}
where $ J(k)=\sum_{j=1}^{k}\sum_{x_{i}\in C_{j}}(x_{i}-\bar{x}_{j})^{2}$, $C_{j}$ is the $j^{th}$ cluster.

Sugar et al.~\cite{jump} proposed jump method which maximizes a modified measure derived from $J(k)$ defined above.
\begin{equation}
\label{eq:jump}
jump(k)=J(k)^{-\frac{p}{2}} - J(k-1)^{-\frac{p}{2}} \\
\text{, where $p$ is the dimension of the dataset}.
\end{equation}

Hartigan et al.~\cite{hartigan} proposed choosing the smallest value of $k$ such that $H( k ) \leq 10$. 
\begin{equation}
\label{eq:hartigan}
H(k)=(\frac{J(k)}{J(k+1)}-1)*(n-k-1)
\end{equation}

In \cite{CH} $k$, which maximizes the following measure,  $CH(k)$, is selected.
\begin{equation}
\label{eq:ch}
CH(k)=\dfrac{(J(1)-J(k))/(k-1)}{J(k)/(n-k)} \\
\end{equation}
Rousseeuw\cite{sil} proposed silhouettes width as described in Eq.\eqref{eq:sil}. It examines how well $i^{th}$ data point in the dataset is clustered.
The term $a(i)$ is the average distance between the $i^{th}$ data point and other data points in its cluster, and $b(i)$ is the lowest average distance between the $i^{th}$ data point to the datapoints of any other cluster of which $i$ is not a part. The value of $s(i)$ lies between -1 to +1. The number of clusters is given by the value, which maximizes the average value of $s(i)$.
\begin{equation}
\label{eq:sil}
s(i)=\dfrac{b(i)-a(i)}{max[a(i),b(i)]} 
\end{equation}
Tibshirani et al.~\cite{gap} proposed a gap statistic, which is described in Eq.\eqref{eq:gap1}
\begin{equation}
\label{eq:gap1}
Gap(k)=\frac{1}{B}\sum_{b}log(J^{*}_{b}(k)) - log(J(k))\\
\end{equation}

This method compared the change of  within-cluster variance with respect to the expected within-cluster variance in a relevant reference null model of the original dataset.
Optimal cluster number is selected when $\log J(k)$ falls below the reference curve. In this method, the range of values of each feature in the original dataset is recorded. Accordingly, reference distribition is obtained by generating each feature uniformly over that range.
$B$ copies of reference dataset are generated.  $J^{*}_{b}(k)$ is the within-cluster variance for the $b^{th}$ uniform dataset. 
Then, standard deviation $s_{d_{k}}$ of $log(J_{b}^{k})$ is computed for $B$ copies of reference dataset and to generate standard deviation at $k$,  $s_{k}$  is computed as $s_{k}=s_{d_{k}}\sqrt{1+\frac{1}{B}}$. 
The estimate of the cluster number is the smallest $k$ such that Eq.\eqref{eq:gap2} holds.

\begin{equation}
\label{eq:gap2}
Gap(k)\geq Gap(k+1)-S_{k+1}\\
\end{equation}
Junhui Wang \cite{CV_A} evaluated the cluster number via cross-validation 
as shown in Eq.\eqref{eq:cv_a}. In this method, the original dataset $X=\{x_{1},\dots,x_{n}\}$ is permuted C times. Each permutation $X^{*c}=\{x_{1}^{*c},\dots,x_{n}^{*c}\}$ is split into 3 sets of size $m, m$ and $n-2m$. First two sets are training data and another set is validation data. $\Psi_{1}^{*c}$ and $\Psi_{2}^{*c}$
model two training datasets. The validation dataset is clustered based on these two models. Instability measure is taken for each $c={1,\dots,C}$ using 
Eq.\eqref{eq:cv_s} and optimum $k$ is evaluated using Eq.\eqref{eq:cv_opt}.
\begin{subequations}
	\label{eq:cv_a}
	
	\begin{equation}
	\label{eq:cv_opt}
	\begin{array}{ll}
	\hat{k}=argmin_{2\leq k \leq K} \hat{s}(\Psi,k,m)&
	\end{array}
	\end{equation}
	
	\begin{equation}
	\label{eq:cv_sa}
	\hat{s}(\Psi,k,m)=\frac{1}{C}\sum_{c=1}^{C}\hat{s}^{*c}(\Psi,k,m)
	\end{equation}
	
	\begin{equation}
	\label{eq:cv_s}
	\hat{s}^{*c}(\Psi,k,m)=\sum_{2m+1\leq i < j \leq n} V_{ij}^{*c}(\Psi,k,z_{1}^{*c},z_{2}^{*c})
	\end{equation}
	
	\begin{equation}
	\label{eq:cv_v}
	V_{ij}^{*c}(\Psi,k,z_{1}^{*c},z_{2}^{*c}) = I[I\{\Psi_{1}^{*c}(x_{i}^{*c}= \Psi_{1}^{*c}(x_{j}^{*c})\} + I\{\Psi_{2}^{*c}(x_{i}^{*c}=\Psi_{2}^{*c}(x_{j}^{*c})\}=1]
	\end{equation}
	
\end{subequations}
$I(.)$ is an indicator function, and $V_{ij}^{*c}(\Psi,k,z_{1}^{*c},z_{2}^{*c})$
measures instability in validation data of $c^{th}$ permutation of the dataset. It counts the number of pairs of elements in validation dataset which are labeled differently by two models $\Psi_{1}$ and $\Psi_{2}$. In this case in each instance, if $\Psi_{1}$ predicts that a pair of data points belong to the same cluster, $\Psi_{2}$ predicts them in different clusters and vice versa. $\hat{s}(\Psi,k,m)$ measures the average instability of the dataset for cluster number $k$.

\subsection{Baseline approaches on k-means Clustering}\label{sub:Bkmeans}
We compare our clustering algorithm  with three state of the art methods. We provide a brief description of them in the following sections.
\subsubsection{Bootstrap Averaging~\cite{Davidson2003}}
This work revolves around the fact that randomly selected large sampled dataset may include a few instances from every class. Therefore, any estimate of sampled dataset lies within some interval of the estimate of the original dataset. In their work, $t$ sampled datasets are made with replacement sampling. \emph{k-means} clustering is applied to each of them. Hence, $t$ sampled datasets of $k$ cluster centroids obtained. They group similar cluster centers of $t$ sampled datasets using a signature which follows Eq.\eqref{eq:signature}. The signature has a range of $0$ to $2^{l+1}$ as each attribute of the dataset is normalized between $0$ to $1$. All ($t \times k$) signatures are sorted in ascending order. Finally, they are divided into $k$ equal intervals to form the groups.

\begin{equation}
\label{eq:signature}
Signature(c_{ij})=\sum_{l}c_{ijl}\times 2^{l}
\end{equation}
where $c_{ijl}$ is the $l$ th attribute for	the $j$ th cluster of the $i$ th sample.\\

\subsubsection{RPKM~\cite{CAPO2017}}
Recursive partition based \emph{k-means} (\emph{RPKM}) is an approximation algorithm whose focus is on reducing distance computations. Cap\'o et al.~\cite{CAPO2017} created partitions of the data space into $d$-dimensional hypercubes using a generic quad tree of $d$-dimensions. They used specific depth $D$ for a given dataset.
Each leaf node at depth $D$ contains  a list of similar data points. Each node in the generic $d$-dimensional quad tree is characterized by two attributes: representative($\bar{S}$) and cardinality($|S|$). They computed representative and cardinality for
all nodes recursively starting from leaf nodes at depth $D$. Initially, the representative for each leaf node is generated by averaging all the points stored in its list and cardinality is set to the total number of datapoints in its list. In the next recursion, nodes at depth $D-1$ is considered. Here, representative and cardinality of each node are computed following Eq.~\ref{eq:rpkm}. In this way, attributes of all nodes in the generic quad tree are computed.\\

\begin{equation}
\begin{array}{lllllll}
\label{eq:rpkm}
|S|&=&\sum_{R \epsilon P_{i+1}[S]}|R|&,&
\bar{S}&=&\frac{\sum_{R \epsilon P_{i+1}[S]}|R|.\bar{R}}{|S|}\\
\end{array}
\end{equation}
Where $|.|$ denotes the cardinality of a node and $\bar{S}$ denotes the representative of a node. $P_{i+1}$ represents children of node $i$.
\\\\Next, \emph{k-means} is executed at each level of generic quad-tree in a bottom-up manner. Representatives of all nodes in depth $i$ are used as data points, and Lloyd's \emph{k-means} is executed on these set of points. If $i$ indicates bottom level, then random $k$ cluster centroids are chosen from all representatives at depth $i$ as initial cluster centroids. Otherwise, $k$ cluster centroids from depth $i+1$ are chosen initially. The recursion stops at the depth where number of nodes is less than $k$. The execution of Lloyd's \emph{k-means} in this depth provides final $k$ cluster centroids for that dataset.
To deal with high dimensionality, authors decided to use at most 8 data attributes to make the quad-tree feasible in any situation. Attributes are randomly selected . In their method, the number of data-points gets reduced at each depth of the quad tree. This also reduces the number of distance computation.

\subsubsection{Refining Seed~\cite{Bradley1998RefiningIP}}
This is also a sampling-based clustering method where sampling is used to find better initial $k$ cluster centroids. It is claimed that a better clustering can be achieved with \emph{k-means} algorithm if initial cluster centroids are appropriately chosen. To refine initial cluster centroids, a heuristic is proposed in this work as follows: \emph{k-means} is executed on each of $J$ sampled dataset with a random selection of initial centroids.  Let this set of centroids be denoted by $CM$. \emph{k-means} is executed on $CM$ using each of its element ($k$ cluster centroids) in $CM$ as initial cluster centroids, and the values of $k$ cluster centroids are stored in $FM$. Finally, the distortion between each item $FM_{i}$ in $FM$ and $CM$ are computed. Distortion is defined as  the sum of squared distances of each data point in $CM$ to its nearest centroid in $FM_{i}$. The item with minimum distortion in $FM$ is selected as refined initial cluster centers. Finally,  \emph{k-means} is executed on the original dataset where refined $k$ cluster centroids are taken as initial $k$ cluster centroids.

\section{Experimental Results}
In this section, we first discuss the prediction of cluster number  with a few comparable methods using synthetic and real-world dataset.
 We used a variety of models in synthetic simulation, including those considered by Tibshirani et al.~\cite{gap} and Dudoit et al.~\cite{Dudoit_02} 
for estimating the number of clusters in a dataset. Finally, we discuss the validity of cluster number on the basis of the clustering outcome. 
%
%
 All data sets were tested in Python 2.7, and were run on an \emph{HP} ProBook
computer with \emph{Intel\textregistered~Core i5 7th Gen}.
\subsection{Quality of clusters}
Here, we describe the metrics, used for inferring clustering results. 
As existing clustering algorithms require a predefined
number of clusters, we set it to the  number, which is either evaluated in the process of execution of the \emph{CNAK} or available from the ground truth.

\subsubsection{Performance Metrics}
For evaluating clustering methods, clustering evaluation metrics, namely adjusted rand index (ARI)~\citep{Hubert1985}, 
normalized mutual information (NMI)~\citep{NMI}, homogeneity (H)~\citep{vMeasure}, completeness (C)~\citep{vMeasure}, and silhouette coefficient~\citep{sil} are used. First four metrics are used to measure the similarity between true labels and model's prediction about data-points. Their upper bound is $1$. Also, the lower bound is $0$ for all the metrics except {\emph ARI}. It's lower bound is not well defined. Any value, less or equal to 0, indicates two random partitions. For a good clustering, higher values are desirable for all the metrics. \emph{ARI} considers how similarly two partitions evaluate a pair of data points. On the other hand, \emph{NMI} is an information theoretic approach which states the quantity of information shared by two partitions. Homogeneity and completeness depict class (true labels) and cluster (predicted labels) containment, i.e., whether all the data points of one cluster belongs to one class and vice versa.
The \emph{silhouette} coefficient describes the quality of the cluster. Its value lies between -1 to +1. A higher value indicates appropriate clustering, i.e., data points within a cluster are very similar, and data points between neighboring clusters are distinct. Lower and negative values indicate too many or too few clusters.

\subsubsection{Stability Metric}
The proposed and comparative algorithms are sampling based, and randomness is associated with initialization of those techniques. Hence, we have executed every method 50 times, for measuring their stability. We provide an average of all metrics \emph{(SM)} computed upon 50 trials and also provide the mode of those metrics in 50 trials. Mode indicates the most frequently occurring value in 50 trials. Therefore, the similarity between \emph{SM} and Mode is an indicator of the stability criterion of our algorithm.

\begin{figure}[htb!]
	
	\centering
	\begin{multicols}{4}
		\subcaptionbox{sim-2\label{fig:sim2}}{\includegraphics[width=0.25\textwidth]{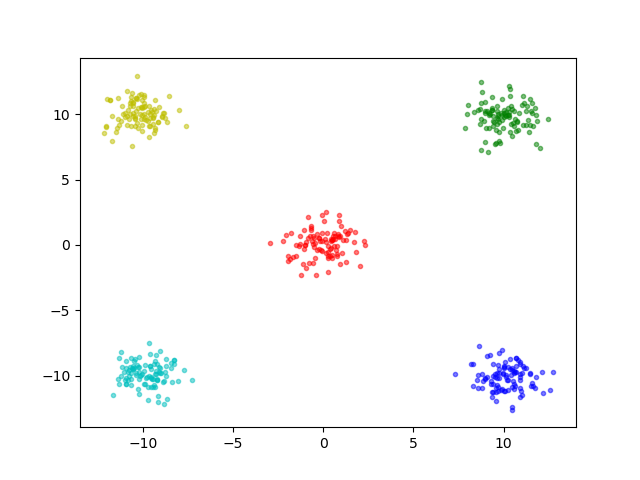}}
		
		\subcaptionbox{sim-3\label{fig:sim3}}{\includegraphics[width=0.25\textwidth]{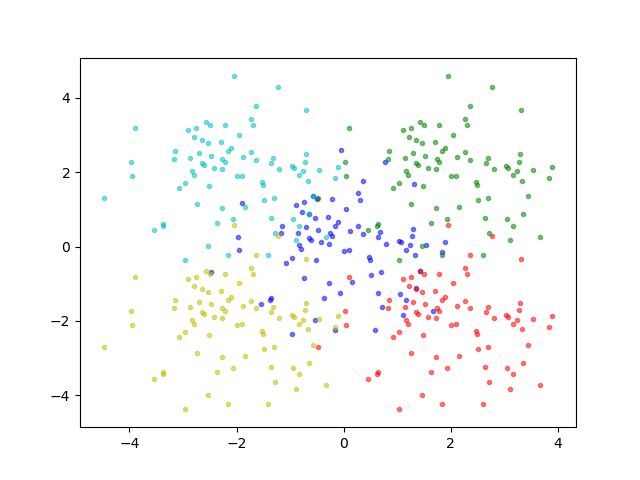}}
		
		\subcaptionbox{sim-4\label{fig:sim4}}{\includegraphics[width=0.25\textwidth]{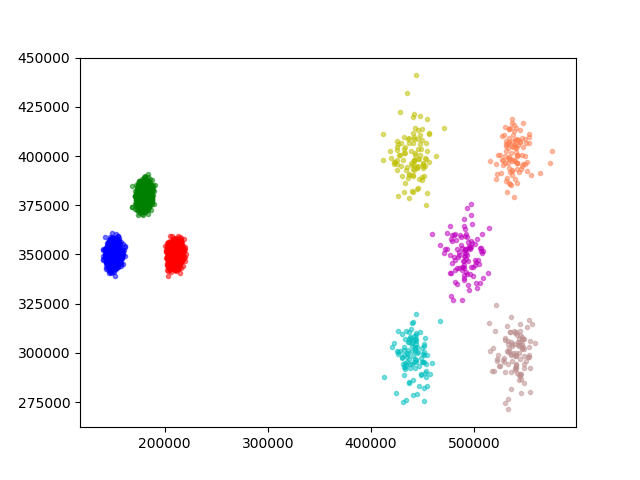}}
		\subcaptionbox{sim-5\label{fig:sim5}}{\includegraphics[width=0.25\textwidth]{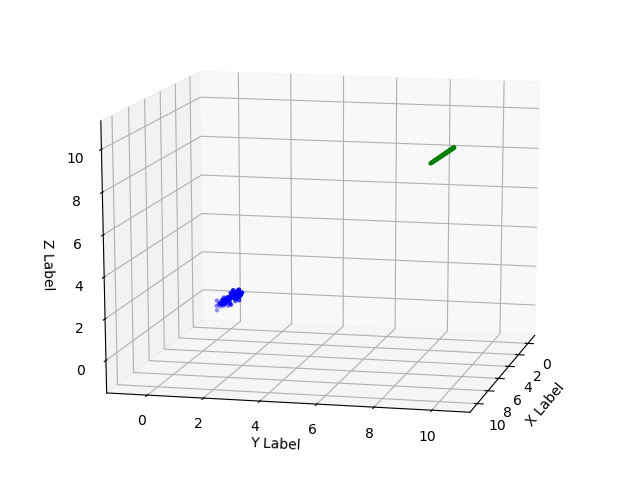}}

	\end{multicols}
	\begin{multicols}{4}
		\subcaptionbox{sim-6\label{fig:sim7}}{\includegraphics[width=0.25\textwidth]{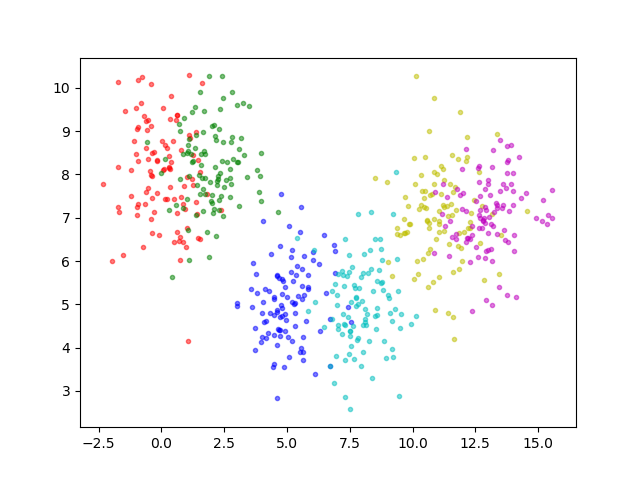}}
		
		\subcaptionbox{sim-7\label{fig:sim8}}{\includegraphics[width=0.25\textwidth]{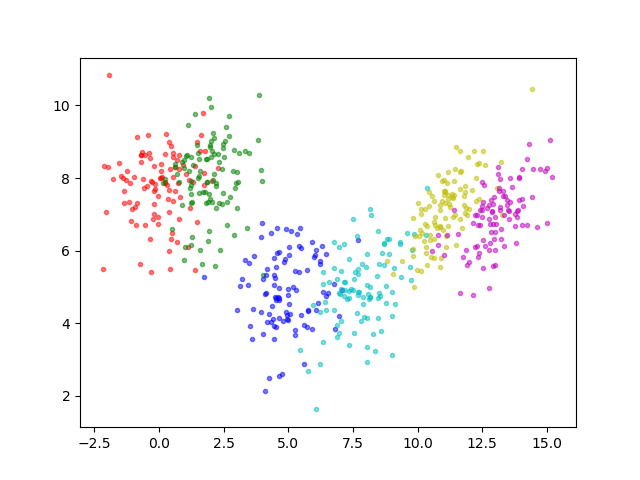}}
		
		\subcaptionbox{sim-8\label{fig:sim9}}{\includegraphics[width=0.25\textwidth]{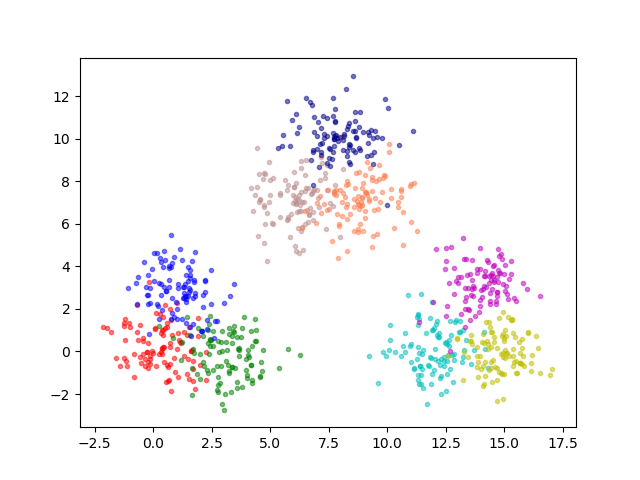}}
		
		\subcaptionbox{sim-9\label{fig:sim10}}{\includegraphics[width=0.25\textwidth]{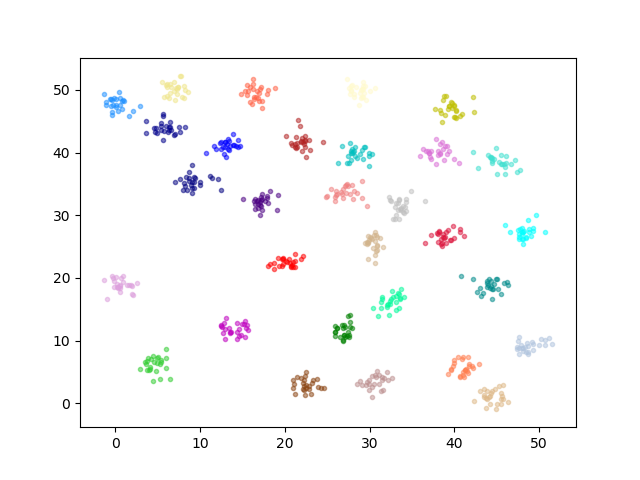}}
		
	\end{multicols}
	\begin{multicols}{4}
		
		
	\end{multicols}

	\caption{Scatter plot of synthetic data simulations where dimension of feature space is $\leq 3$.}
	\label{fig:scatterplot}
\end{figure}
\subsection{Synthetic Dataset}
We describe simulations into two parts.  Each simulation contains a dataset in the first part, whereas the other part contains a group of datasets to reveal some characteristics of clustering. We name the first part as ``Singleton Simulation" \emph{(SS)} and the second part  as ``Group Simulation" \emph{(GS)}. \emph{SS} includes synthetic datasets that include the following criterion: 1) single cluster, 2) clusters with varying ``\emph{between-cluster separation}", 3)impact of clusters with imbalance size, 4) clustering of a large number (more than a million) of instances, 5) feature spaces with different types of a probability distribution, 6) dependency among features, and 7) different degrees of overlap. 
On the other hand, \emph{GS} examine the following challenges: 1) robustness of a clustering method against the extent of overlapping of two clusters, 
2) clustering high dimensional dataset, and 3) detection of any shaped clusters. We use ``Numpy" package in ``Python 2.7" to create these simulations.

\subsubsection{Description of synthetic simulation}\label{sec:synsim}
We have created twelve simulations considering afore-mentioned characteristics. 
\begin{enumerate}
	\item \emph{sim-1:} One cluster in a twenty dimensional $(R^{20})$ feature space. This contains 200 observations with a uniform distribution over the unit hypercube in a 10-dimensional space.
	
	\item \emph{sim-2:} Five clusters in 2-D (two-dimensional $(R^{2})$ real space). The observations in each of the five clusters are independent bivariate normal random variables with means (0, 0), (10, 10), (10, -10), (-10, 10), and (-10, -10), respectively, and identity covariance matrix. There are 100 observations in each of the 5 clusters, respectively.

	\item \emph{sim-3:} Five clusters in $R^2$. The observations in each of the five clusters are independent bivariate normal random variables with means $(0, 0), (2, 2), (2,-2), (-2, 2),$ and $(-2. -2)$, respectively, and identity covariance matrix. There are 100 observations in each of the 5 clusters.
	
	\item \emph{sim-4:} Eight clusters in $R^2$. Five clusters have 100 observations each, and three clusters have  2000 observations each. The observations in each of the clusters are independent bivariate normal random variables with means are similar to~\cite{UnbalanceSet} and identity covariance matrix.
	
	
	\item \emph{sim-5:} Two elongated clusters in $R^3$. Cluster $1$ contains $100$ observations generated such that each dimension is taking on equally spaced values from $-0.5$ to $0.5$. Gaussian noise with a standard deviation of $0.1$ and $0$ mean is then added to each variable. Cluster $2$ is generated in the same way except that the value $10$ is added to each variable instead of Gaussian noise. This results in two elongated clusters, stretching out along the main diagonal of a three-dimensional cube, with $100$ observations each.

	
	\item \emph{sim-6:} Six clusters in $R^2$. The observations in each of the six clusters are independent bivariate normal random variables with means (0, 8), (2, 8), (5, 5), (8, 5), (11, 7), and (13, 7), respectively, and identity covariance matrix. There are $100$ observations in each of the $6$ clusters, respectively. 
	This simulation is generated to explore cluster overlapping. The distance between centroids of two overlapping clusters is in between 2 and 3.
	
	\item \emph{sim-7:} Six clusters in $R^{2}$. This simulation is generated as in \emph{sim-6}. But, we modify the covariance structure. The correlations for six clusters are 0.0, 0.0, 0.3, 0.3, 0.7 and 0.7 corresponding to cluster mean in \emph{sim-6}.
	\item \emph{sim-8:} Nine clusters in $R^{2}$. The observations in each of the nine clusters are independent bivariate normal random variables with means (0, 0), (3, 0), (1, 3), (12, 0), (15, 0), (14, 3), (6,7), (9,7), and (8,10), respectively, and identity covariance matrix. There are $100$ observations in each of the $9$ clusters, respectively. This simulation is simulated with the same purpose of {sim-6}. But, we made ``between-cluster separation" among three groups very small. Also, these groups with three closely space clusters are far from each other. 
	
	\item \emph{sim-9:} Thirty clusters in $R^{2}$. Each cluster contains 25 observations. They are an independent bivariate normal random variable with identity covariance matrix and appropriate mean vector. Each mean is randomly generated and  has a range between 0 to 50. Any simulation where the {\emph Euclidean} distance between the mean of two clusters is less than 1, is discarded.

	
	\item \emph{sim-10:} Four clusters in $R^{4}$. The observations in each of the four clusters are independent multivariate normal random variables with means (1, 10, 30, 50), (10, 30, 50, 70), (30, 50, 70, 90), and  (50, 70, 90, 110), respectively, and identity covariance matrices. There are $50500$ observations in each of the four clusters.
	
	
	\item \emph{sim-11:} Four clusters in $R^{5}$. The observations in each of the four clusters are independent multivariate normal random variables with means 	(100, 50, 0, 150, 150), (150, 100, 50, 0,  200), (50, 0, 150, 200, 100), and  (0, 200, 100, 50, 150), respectively, and identity covariance matrices. There are $2\times10^{6}$ observations in each of the 5 clusters.
	
	
\end{enumerate}

\begin{table}[!htb!]
	\caption{Determination of the size of a sampled dataset with its parameters }
	\begin{tabular}{lcccc|lcccc}
		\hline
		Dataset & \multicolumn{1}{c}{$\lambda_{max}$} & \multicolumn{1}{c}{$\tau$} & \multicolumn{1}{l}{$c$} & \multicolumn{1}{c}{\begin{tabular}[c]{@{}l@{}}Sample \\ size(\%)\end{tabular}} & \multicolumn{1}{|l}{Dataset} & \multicolumn{1}{c}{$\lambda_{max}$} & \multicolumn{1}{l}{$\tau$} & \multicolumn{1}{c}{$c$} & \multicolumn{1}{c}{\begin{tabular}[c]{@{}l@{}}Sample \\ size(\%)\end{tabular}} \\
		\hline
		sim-1 & 0.079 & - & 0.2 & 4 & sim-7 & 22.62 & - & 0.6 & 28.72 \\
		sim-2 & 81.20 & 16 & 1.32 & 31 & sim-8 & 36.20 & - & 0.6 & 34\\
		sim-3 & 4.32 & - & 0.20 & 50 &  sim-9 & 402.88 & 16 & 1.45 & 7\\
		sim-4 & 19840250263 & 2 & 1.00 & 100 & sim-10 & 1915.54 & 8 & 2.57 & 0.0029 \\
		sim-5 & 75.64 & - & 0.6 & 80 & sim-11 & 10435.27 & 8 & 3.179 & 0.000496 \\
		sim-6 & 20.05 & - & 0.6 & 60  & &&&&\\\hline
	\end{tabular}
\label{tab:SS_size}
\end{table}

\subsubsection{Prediction of cluster number in \emph{SS}}
\begin{figure}[htb!]
	
	\centering
	\begin{multicols}{4}
		\subcaptionbox{sim-1\label{fig:plot_sim1}}{\includegraphics[width=0.25\textwidth]{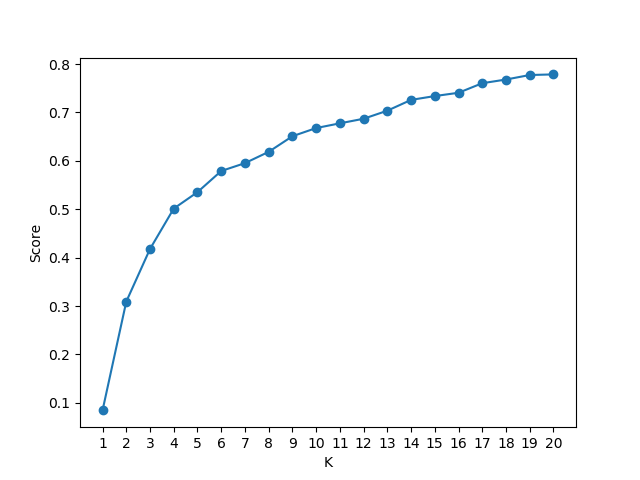}}
		
		\subcaptionbox{sim-2\label{fig:plot_sim2}}{\includegraphics[width=0.25\textwidth]{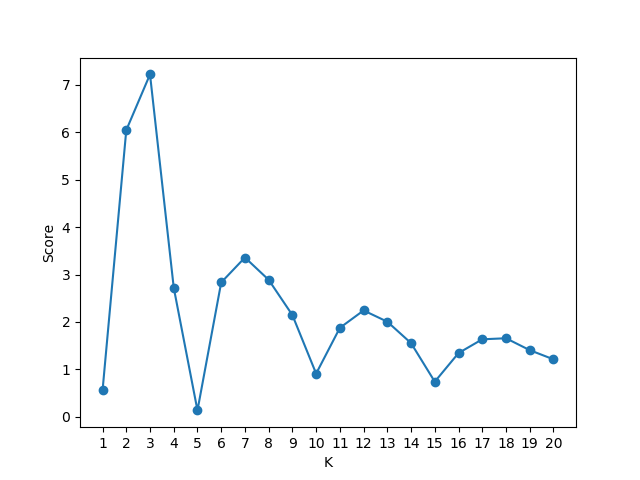}}
		
		\subcaptionbox{sim-3\label{fig:plot_sim3}}{\includegraphics[width=0.25\textwidth]{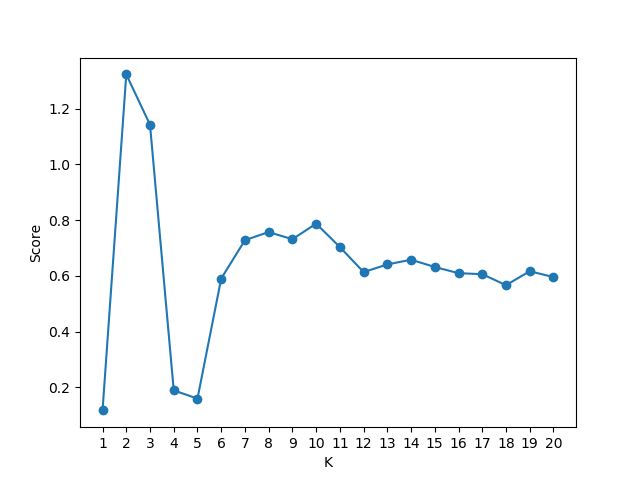}}
		\subcaptionbox{sim-4\label{fig:plot_sim4}}{\includegraphics[width=0.25\textwidth]{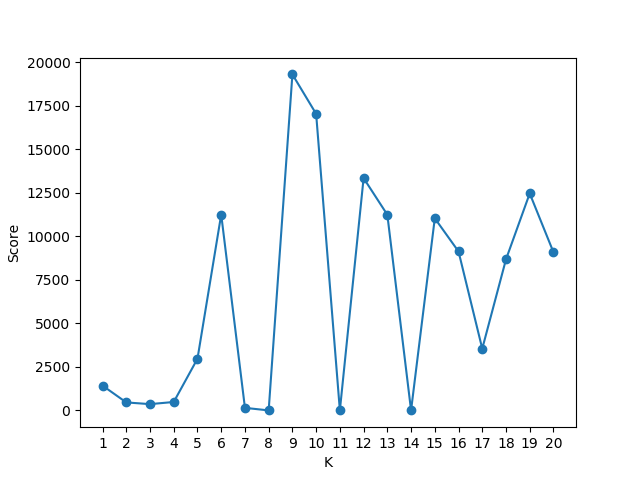}}
				
	\end{multicols}
	\begin{multicols}{4}
		\subcaptionbox{sim-5\label{fig:plot_sim5}}{\includegraphics[width=0.25\textwidth]{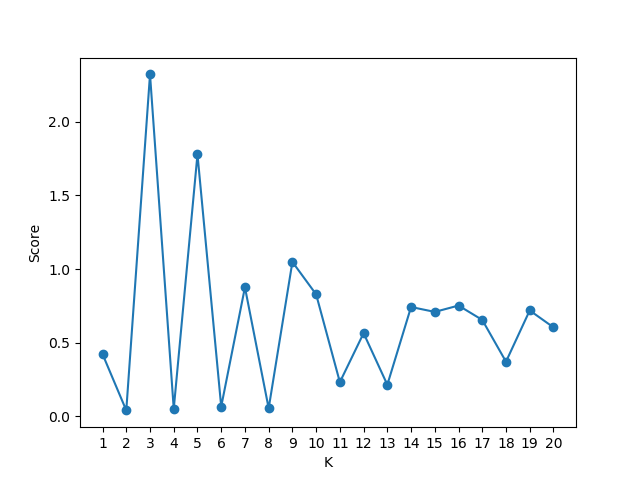}}

		
		\subcaptionbox{sim-6\label{fig:plot_sim7}}{\includegraphics[width=0.25\textwidth]{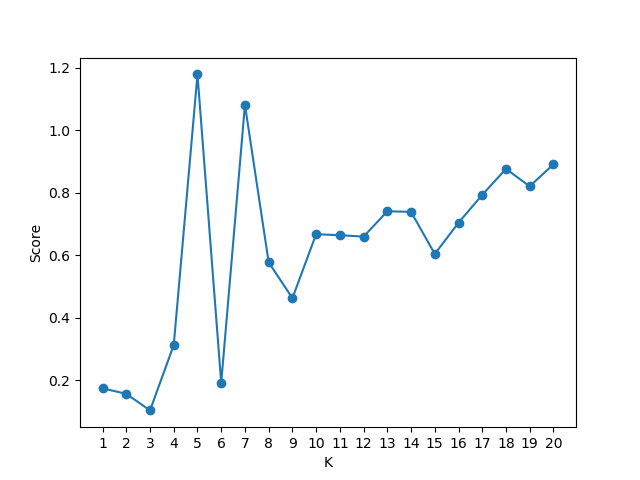}}
		
		\subcaptionbox{sim-7\label{fig:plot_sim8}}{\includegraphics[width=0.25\textwidth]{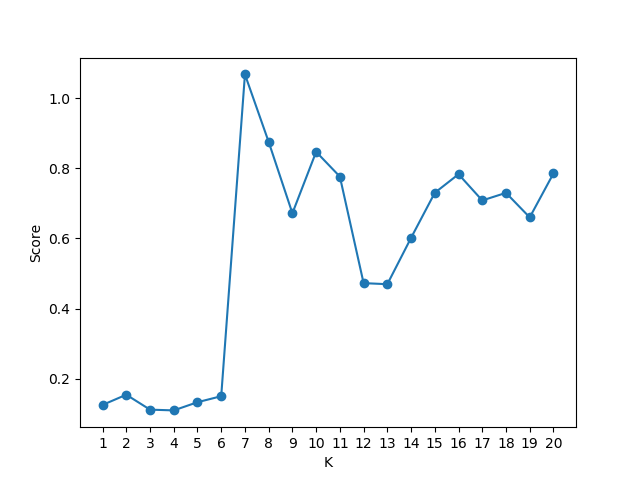}}

		\subcaptionbox{sim-8\label{fig:plot_sim9}}{\includegraphics[width=0.25\textwidth]{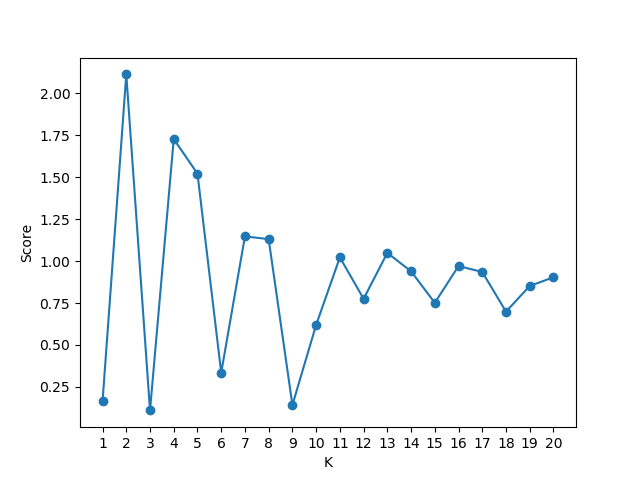}}
		
	\end{multicols}
	\begin{multicols}{4}

		\subcaptionbox{sim-9\label{fig:plot_sim10}}{\includegraphics[width=0.25\textwidth]{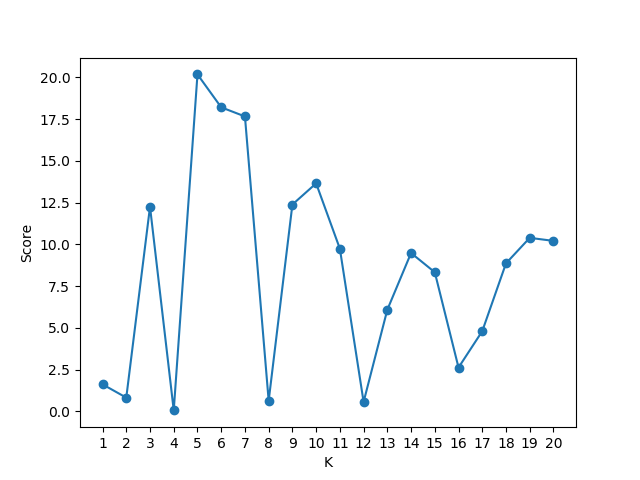}}
		
		\subcaptionbox{sim-10\label{fig:plot_sim11}}{\includegraphics[width=0.25\textwidth]{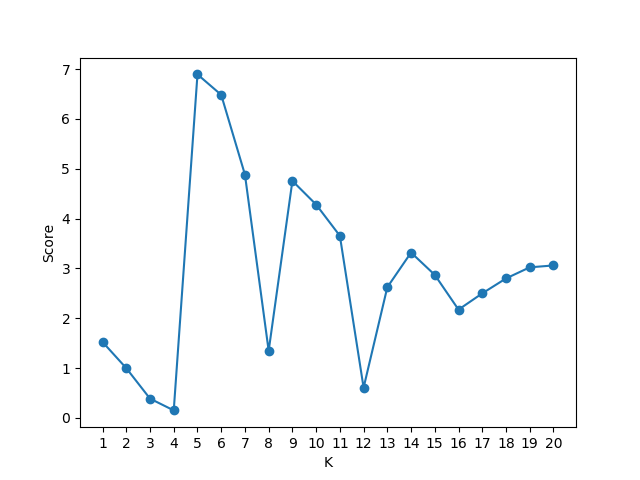}}
		
		\subcaptionbox{sim-11\label{fig:plot_sim12}}{\includegraphics[width=0.25\textwidth]{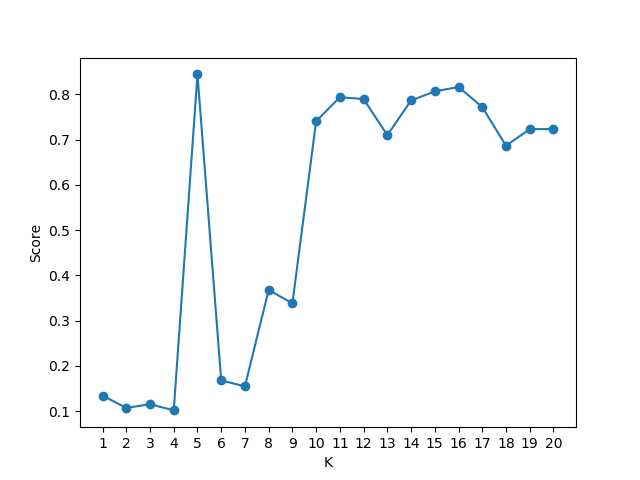}}

	\end{multicols}

	\caption{plot of ``k vs Score". Score is generated by \emph{CNAK}}
	\label{fig:plot_k_score}
\end{figure}

\begin{figure}[htb!]	
	\centering
	\includegraphics[width=\textwidth]{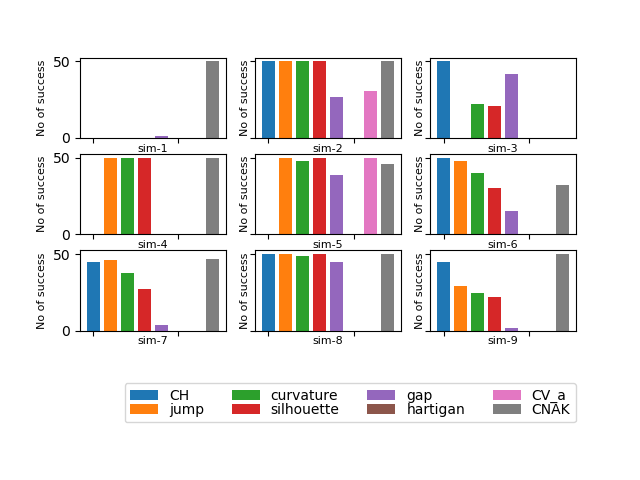}
	\caption{Prediction of cluster number by eight methods. For each of the ten simulations (in \emph{SS}), the bar plots represent the number of successful attempts (out of 50 realizations) where true cluster number is predicted.}
	\label{fig:bar_SS}
\end{figure}

The proposed method \emph{CNAK} compared with seven existing
methods presented in Section~\ref{sub:BCN}. We have generated 50 realizations for each of the simulation in \emph{SS}.
We have shown the outcome of the methods in Table~\ref{tab:SS}. We put ``*" in the cell where the prediction is disseminated over many values between $k_{min}$ to $k_{max}$ for 50 realizations. Hence, conclusion can not be drawn. We put ``-" where execution time for a single realization is very high, and use ``$\times$" where methods are not applicable. We follow this throughout our paper.
Results on \emph{sim-1} suggest that \emph{CNAK} can predict a single cluster. Therefore, it can indicate whether clustering is required at all. Though the \emph{gap} is another popular method in the literature which can predict such a situation, it fails to perform in our simulation (\emph{sim-1}). However, \emph{CNAK} indicates a single cluster for \emph{sim-3}. We made  ``between-cluster separation" very small for \emph{sim-3} (shown in Figure~\ref{fig:sim3}). Therefore, it appears as a single cluster. This situation can be circumvented by looking at the plot of ``k vs. score" with detailed analysis. Figure~\ref{fig:plot_sim3} shows another dip at $k=5$, which is the second minimum score, computed by the \emph{CNAK}. The experimental simulation shows that \emph{CNAK} is good at identifying well-separated clusters (\emph{sim-3} and \emph{sim-5}) as the other methods under comparison. 
We also study the robustness of prediction techniques in the presence of cluster imbalance.  `Cluster imbalance' is a phenomenon where the number of instances for representing a few groups is small compared to the other groups.
Results indicate that the \emph{CNAK} along with the \emph{curvature, silhouette}, and \emph{jump} methods can take care of cluster imbalance.
Another advantage of our approach is to estimate cluster number for a large dataset which contains more than a million instances. \emph{CNAK} takes only 34.57 seconds and 45 minutes for executing \emph{sim-10} and \emph{sim-11}, respectively. On the other hand, other comparing methods take hours to days. 
To justify the performance of \emph{CNAK}, we generate a score ($gs$) using Eq.~\ref{eq:stat}.

\begin{equation}
\label{eq:stat}
\begin{array}{lll}
gs&=&\exp^{-(\dfrac{k_{True}-k_{X}}{\sigma})^{2}}
\end{array}
\end{equation}
Where $k_{True}$ is the ground truth of the dataset, and $k_{X}$ is the predicted cluster number by method $X$. We set $\sigma=2$ for our experiment. The value of gs lies between 0 to 1. Higher its value, better is the estimation. 
We consider \emph{gs-score} to be 0 for sim-1, sim-10, and sim-11 if any method is not able to compute their score for $k=1$, or is not possible to execute within a reasonable execution time.  
To summarize, we have shown a bar graph for true $k$ in Figure~\ref{fig:bar_SS},  which indicates the consistent and comparable behavior of \emph{CNAK}.
Average \emph{gs} score in Table~\ref{tab:SS} indicates consistent behavior of the \emph{CNAK} for all simulations under consideration. We have highlighted the best performances in bold font. The novel contributions in our approach are correct identification of the single cluster $k=1$, and fast computation for estimating the number of clusters in big data.

\begin{table}[htb!]
	\caption{Prediction of cluster number based on the mode of accumulated estimation of 50 realizations.}
	\label{tab:SS}
	\begin{tabular}{lccccccccc}
		\hline
		\multirow{2}{*}{Dataset} & \multicolumn{1}{l}{\multirow{2}{*}{\begin{tabular}[c]{@{}l@{}}True\\   K\end{tabular}}} & \multicolumn{8}{c}{Comparison of seven approached in cluster number prediction} \\ \cline{3-10} 
		& \multicolumn{1}{l}{} & CH & Jump & Hartigan & Curvature & Silhouette & Gap & $CV_{a}$ & CNAK \\ \hline
		sim-1 & 1 & $\times$ & $\times$ & $\times$ &$\times$& $\times$ & $*$ & $\times$ & 1 \\
		sim-2 & 5 		& 5 	& 5 & * & 5 & 5 & 5 & 5 & 5 \\
		sim-3 & 5 		& 5 	& 4 & * & 5 & 4 & 5 & 4 & 1 \\
		sim-4 & 8 		& 14 	& 8 & * & 8 & 8 & 17 & 7 & 8 \\ \hline
		sim-5 & 2 		& 20 	& 2 & 9 & 2 & 2 & 2 & 2 & 2 \\
		sim-6 & 3,6 	& 3 	& 3 & * & 3 & 3 & 3 & 2 & 3 \\
		sim-7 & 3,6 	& 3 	& 3 & * & * & 3 & 4 & 2 & 3 \\ \hline
		sim-8 & 3,9 	& 3 	& 3 & * & 3 & 3 & 3 & 3 & 3 \\
		sim-9 & 30 	& 30 	& 30& 32& 30& 29,30 & 35 & 2 & 30 \\
		sim-10 & 4 & - & - & - & - & - & - & - & 4 \\
		sim-11 & 4 & - & - & - & - & - & - & - & 4\\\hline
		\multicolumn{2}{c}{Average gs-score}&0.524  & 0.706 & 0.000 & 0.564 & 0.687 & 0.596 & 0.626 & {\bf 0.904}\\ \hline
	\end{tabular}
\end{table}

\subsubsection{Clustering on Synthetic Dataset (SS)}

\begin{table*}[htb!]
	\centering
	\caption{Clustering results on the synthetic simulations (sim2-sim7)}
	\makebox[\linewidth]{
		\begin{tabular}{llccccccccccc}
			\hline
			\multirow{2}{*}{Dataset} & \multirow{2}{*}{Methods} & \multirow{2}{*}{K} & \multicolumn{2}{c}{ARI}  & \multicolumn{2}{c}{NMI} & \multicolumn{2}{c}{Homogeneity} & \multicolumn{2}{c}{Completeness} & \multicolumn{2}{c}{Silhouette} \\ \cline{4-13} 
			&  &  & Mode & SM & Mode & SM & Mode & SM & Mode & SM & Mode & SM \\ \hline

			\multirow{4}{*}{\emph{sim-2}} &{\bf CNAK} & 5 & 1.00 & 1.00 & 1.00 & 1.00 & 1.00 & 1.00 & 1.00 & 1.00 & 0.91 & 0.91 \\ 
			&{\bf BA} & 5 & 1.00 & 1.00 & 1.00 & 1.00 & 1.00 & 1.00 & 1.00 & 1.00 & 0.91 & 0.91 \\ 
			&{\bf RS} & 5 & 1.00 & 1.00 & 1.00 & 1.00 & 1.00 & 1.00 & 1.00 & 1.00 & 0.91 & 0.91 \\ 
			& RPKM & 5 & 0.78 & 0.83 & 0.91 & 0.93 & 0.83 & 0.87 & 0.99 & 0.99 & 0.51 & 0.36 \\ \hline

			\multirow{4}{*}{\emph{sim-3}} & \multirow{1}{*}{\bf CNAK} 
			& 5 & {\bf 0.68} & 0.64 &{\bf 0.69} & 0.66 &{\bf 0.69} & 0.57 & {\bf 0.69} & 0.66 & {\bf 0.60} & {\bf 0.59} \\ 
			& \multirow{1}{*}{BA} 
			& 5 & 0.63 & 0.63 & 0.64 & 0.63 & 0.63 & 0.63 & 0.64 & 0.64 & 0.56 & 0.56 \\  
			& \multirow{1}{*}{\bf RS} 
			& 5 & 0.66 &{\bf 0.66} & 0.67 &{\bf 0.68} & 0.67 & {\bf 0.68} & 0.68 & {\bf 0.67} & 0.54 & 0.54 \\ 
			& \multirow{1}{*}{RPKM} 
			& 5 & 0.55 & 0.63 & 0.60 & 0.63 & 0.58 & 0.63 & 0.61 & 0.64 & 0.51 & 0.41 \\ \hline

			\multirow{4}{*}{sim-4} &{\bf CNAK} & 8 & 1.00 & 1.00 & 1.00 & 1.00 & 1.00 & 1.00 & 1.00 & 1.00 & 0.99 & 0.99 \\
			&{\bf BA} & 8 & 1.00 & 1.00 & 1.00 & 1.00 & 1.00 & 1.00 & 1.00 & 1.00 & 0.99 & 0.99 \\
			&{\bf RS} & 8 & 1.00 & 1.00 & 1.00 & 1.00 & 1.00 & 1.00 & 1.00 & 1.00 & 0.99 & 0.99 \\
			& {\bf RPKM} & 8 & 1.00 & 1.00 & 1.00 & 1.00 & 1.00 & 1.00 & 1.00 & 1.00 & 0.99 & 0.99 \\\hline

			\multirow{4}{*}{sim-5} & {\bf CNAK} & 2 & 1.00 & 1.00 & 1.00 & 1.00 & 1.00 & 1.00 & 1.00 & 1.00 & 0.96 & 0.96 \\
			&{\bf BA} & 2 & 1.00 & 1.00 & 1.00 & 1.00 & 1.00 & 1.00 & 1.00 & 1.00 & 0.96 & 0.96 \\
			&{\bf RS} &2&1.00 & 1.00 & 1.00 & 1.00 & 1.00 & 1.00 & 1.00 & 1.00& 0.96 & 0.96 \\
			&{\bf RPKM} & 2 &1.00 & 1.00 & 1.00 & 1.00 & 1.00 & 1.00 & 1.00 & 1.00& 0.96 & 0.96  \\\hline


			\multirow{8}{*}{\emph{sim-6}} & \multirow{2}{*}{\bf CNAK} & 3 &{\bf 0.88} &{\bf 0.88} &{\bf 0.83} &{\bf 0.83 }&{\bf 0.83} &{\bf 0.83} &{\bf 0.83} &{\bf 0.83 }& 0.79 & 0.78 \\  
			&  & 6 & 0.47 & 0.47 & 0.64 & 0.64 & 0.81 & 0.82 & 0.50 & 0.50&{\bf 0.58} & {\bf 0.59} \\ 
			& \multirow{2}{*}{BA} & 3 & 0.87 & 0.87 & 0.82 & 0.83 & 0.82 & 0.83 & 0.82 & 0.83 & 0.79 & 0.79 \\  
			&  & 6 & 0.53 & 0.52 & 0.66 & 0.64 & 0.64 & 0.62 & 0.64 & 0.64 & 0.23 & 0.243 \\ 
			& \multirow{2}{*}{\bf RS} & 3 & {\bf 0.88} &{\bf 0.88} &{\bf 0.83} &{\bf 0.83} &{\bf 0.83} &{\bf 0.83} &{\bf 0.83} &{\bf 0.83} &{\bf 0.80} &{\bf 0.80} \\  
			&  & 6 & {\bf 0.63} &{\bf 0.64} &{\bf 0.72} &{\bf 0.72} &{\bf 0.67} & {\bf 0.68} &{\bf 0.67} &{\bf 0.68} & 0.53 & 0.52 \\ 
			& \multirow{2}{*}{RPKM} & 3 & 0.64 & 0.64 & 0.64 & 0.64 & 0.63 & 0.63 & 0.64 & 0.64 & 0.39 & 0.39 \\  
			&  & 6 & 0.66 & 0.61 & 0.71 & 0.69 & 0.71 & 0.67 & 0.71 & 0.71 & 0.40 & 0.37 \\ \hline

			\multirow{8}{*}{sim-7}& \multirow{2}{*}{\bf CNAK}  &3 & {\bf 0.93} &{\bf 0.93} &{\bf 0.90} &{\bf 0.90} &{\bf 0.90} &{\bf 0.90} &{\bf 0.90} &{\bf 0.90} & 0.57 & 0.57 \\
			&  & 6 & {\bf 0.65} &{\bf 0.65} &{\bf 0.73} &{\bf 0.73} &{\bf 0.73} &{\bf 0.73} &{\bf 0.73} &{\bf 0.73} &{\bf 0.40} &{\bf 0.40} \\
			
			& \multirow{2}{*}{BA} & 3 & 0.85 & 0.83 & 0.82 & 0.80 & 0.82 & 0.79 & 0.82 & 0.80 & 0.55 & 0.53 \\
			&  & 6 & 0.61 & 0.61 & 0.71 & 0.69 & 0.71 & 0.70 & 0.70 & 0.69 & 0.30 & 0.30 \\
			
			& \multirow{2}{*}{RS} & 3 & 0.52 & 0.52 & 0.70 & 0.70 & 0.55 & 0.55 & 0.89 & 0.89 & 0.58 & 0.58 \\
			&  & 6 & 0.62 & 0.61 & 0.71 & 0.69 & 0.71 & 0.69 & 0.71 & 0.69 & 0.31 & 0.30 \\
			& \multirow{2}{*}{RPKM} & 3 &0.61 &0.52 & 0.55 & 0.58& 0.58 & 0.53& 0.66  &0.65 & 0.53& 0.48 \\
			& &6 & 0.44 &0.48 & 0.60 &0.59 & 0.53 &0.51 & 0.68 &0.69 &0.34 & 0.30 \\ \hline

		\end{tabular}
	}
	\label{tab:clustering_syn1}
\end{table*}

\begin{table*}[htb!]
	\centering
	\caption{Clustering results on the synthetic simulations (sim8-sim12)}
	\makebox[\linewidth]{
		\begin{tabular}{llccccccccccc}
			\hline
			\multirow{2}{*}{Dataset} & \multirow{2}{*}{Methods} & \multirow{2}{*}{K} & \multicolumn{2}{c}{ARI}  & \multicolumn{2}{c}{NMI} & \multicolumn{2}{c}{Homogeneity} & \multicolumn{2}{c}{Completeness} & \multicolumn{2}{c}{Silhouette} \\ \cline{4-13} 
			&  &  & Mode & SM & Mode & SM & Mode & SM & Mode & SM & Mode & SM \\ \hline

			%

			\multirow{8}{*}{\emph{sim-8}} & \multirow{2}{*}{\bf CNAK} & 3 & {\bf 1.00} &{\bf 1.00} &{\bf 1.00} &{\bf 1.00} &{\bf 1.00 }&{\bf 1.00} &{\bf 1.00} & {\bf 1.00} &{\bf 0.99} &{\bf 0.99} \\  
			&  & 9 &{\bf 0.61} &{\bf 0.58} &{\bf 0.72} &{\bf 0.71} &{\bf 0.72 }&{\bf 0.70} &{\bf 0.72} &{\bf 0.71} &{\bf 0.39} &{\bf 0.38} \\ 
			& \multirow{2}{*}{BA} & 3 & 1.00 & 1.00 & 1.00 & 1.00 & 1.00 & 1.00 & 1.00 & 1.00 & 0.94 & 0.94 \\  
			&  & 9 & 0.54 & 0.51 & 0.68 & 0.67 & 0.68 & 0.67 & 0.69 & 0.67 & 0.25 & 0.22 \\ 
			& \multirow{2}{*}{RS} & 3 & 1.00 & 1.00 & 1.00 & 1.00 & 1.00 & 1.00 & 1.00 & 1.00 & 0.94 & 0.94 \\  
			&  & 9 & 0.57 & 0.57 & 0.71 & 0.71 & 0.70 & 0.71 & 0.72 & 0.71 & 0.28 & 0.29 \\ 
			& \multirow{2}{*}{RPKM} & 3 & 1.00 & 0.98 & 1.00 & 0.99 & 1.00 & 0.98 & 1.00 & 0.99 & 0.76 & 0.75 \\  
			&  & 9 & 0.53 & 0.51 & 0.69 & 0.69 & 0.64 & 0.64 & 0.74 & 0.74 & 0.30 & 0.35 \\ \hline
			
			\multirow{4}{*}{sim-9} & {\bf CNAK} & 30 &{\bf 0.99} &{\bf 0.99} &{\bf 0.99} & {\bf 0.99} &{\bf 0.99} &{\bf 0.99} &{\bf 0.99} &{\bf 0.99} &{\bf 0.75} & {\bf 0.75} \\
			
			& BA & 30 & 0.38 & 0.39 & 0.71 & 0.71 & 0.70 & 0.70 & 0.71 & 0.71 & -0.17 & -0.17 \\
			
			& RS & 30 & 0.77 & 0.77 & 0.94 & 0.94 & 0.90 & 0.90 & 0.94 & 0.94 &0.64  &0.63  \\
			
			& RPKM & 30 & 0.69 & 0.69 & 0.88 & 0.89 & 0.87 & 0.87 & 0.92 & 0.91 & 0.49 & 0.48 \\\hline

			\multirow{4}{*}{\emph{sim-10}} & \multirow{1}{*}{\bf CNAK} & 4 &{\bf 0.97} &{\bf 0.97} &{\bf 0.96} &{\bf 0.96} &{\bf 0.96} &{\bf 0.96} &{\bf 0.96} & {\bf 0.96} & - & - \\  
			& \multirow{1}{*}{BA} & 4 & 0.96 & 0.93 & 0.94 & 0.92 & 0.94 & 0.92 & 0.94 & 0.92 & - & - \\  
			& \multirow{1}{*}{RS} & 4 & {\bf 0.97} & 0.96 & {\bf 0.96} & 0.95 & {\bf 0.96} & 0.95 & {\bf 0.96}& 0.95 & - & - \\  
			& \multirow{1}{*}{RPKM} & 4 & 0.47 & 0.46 & 0.62 & 0.62 & 0.56 & 0.53 & 0.70 & 0.73 & - & - \\ \hline 

			\multirow{4}{*}{\emph{sim-11}} & \multirow{1}{*}{\bf CNAK} & 4 & {\bf 1.00} &{\bf 1.00} &{\bf 1.00} &{\bf 1.00} &{\bf 1.00} &{\bf 1.00} &{\bf 1.00} & {\bf 1.00} & - & - \\  
			& \multirow{1}{*}{BA} & 4 &{\bf 1.00} & 0.94 &{\bf 1.00} & 0.97 & {\bf 1.00} & 0.95 &{\bf 1.00} & 0.99 & - & - \\  
			& \multirow{1}{*}{RS} & 4 & 0.71 & 0.72 & 0.87 & 0.86 & 0.75 & 0.76 & 0.99 & 0.99 & - & - \\  
			& \multirow{1}{*}{RPKM} & 4 & 0.71 & 0.65 & 0.87 & 0.80 & 0.80 & 0.69 & 0.99 & 0.94 & - & - \\\hline  
		\end{tabular}
	}
	\label{tab:clustering_syn2}
\end{table*}

By evaluating the quality of clustering, we may judge and validate the predicted cluster number.  Table~\ref{tab:clustering_syn1} and ~\ref{tab:clustering_syn2} suggest whether our proposed method produces a good clustering with its predicted cluster number. We have designed a pair of simulations to examine the performance of methods in the context of  ``between-cluster separation", ``tolerance against noise variables", ``correlation between feature variables", ``cluster hierarchy" and ``large scale dataset".
\emph{Sim-2} and \emph{sim-3} reveal that the performance of \emph{CNAK} deteriorates with decreasing ``between-cluster separation". 
The \emph{CNAK} can detect the presence of hierarchical clustering structure. But, its performance in clustering declines at an individual level. It is to be noted that a few closely spaced clusters form a big cluster. This is how we define the hierarchy in the simulation. This might be the reason of performance drop at an individual level, as suggested by the scores of k=3 and 6 for \emph{sim-6} in Table~\ref{tab:clustering_syn1}. Also, we observe that clustering improves in \emph{sim-7}. Here, we add a positive correlation between two variables in the covariance matrix while generating this simulation.
Finally, we have seen that \emph{CNAK} and \emph{bootstrap-averaging} techniques produce similar value for clustering outcome except for \emph{sim-9}. This shows  that our proposed method is robust against a large cluster number. Due to a large number of instances (\emph{sim-10} and \emph{sim-11}), we are not able to compute the {\emph silhouette} score. However, the advantage of our proposed method is a large reduction in execution time without compromising accuracy. Also, we designed two single simulations to measure the performance in the case of ``unbalanced clustering" and the presence of ``many clusters". Like other comparable methods, \emph{CNAK} can handle unbalanced clustering situation as suggested by the performance measures for \emph{sim-4}. However, with many clusters, our method performs the best among all other comparable methods. We have highlighted the best performing clustering algorithm and the values of its performance measures in Tables~\ref{tab:clustering_syn1}, and ~\ref{tab:clustering_syn2}. In summary, the \emph{CNAK} is able to detect whether a  dataset forms a single cluster. It is also capable of determining a large number of clusters, efficient in large scale dataset. But, its performance declines with decreasing ``between-cluster separation" 
as expected like any other k-means clustering based method.

\subsubsection{Study on overlapping clusters (sim-g1)}

\begin{figure}[htb!]
	
	\centering
	\begin{multicols}{4}
		\subcaptionbox{dist-6\label{fig:CI-dist6}}{\includegraphics[width=0.25\textwidth]{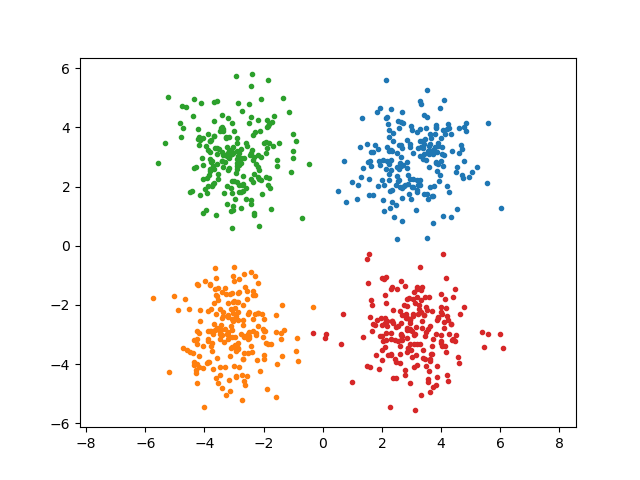}}
		\subcaptionbox{dist-4\label{fig:CI-dist4}}{\includegraphics[width=0.25\textwidth]{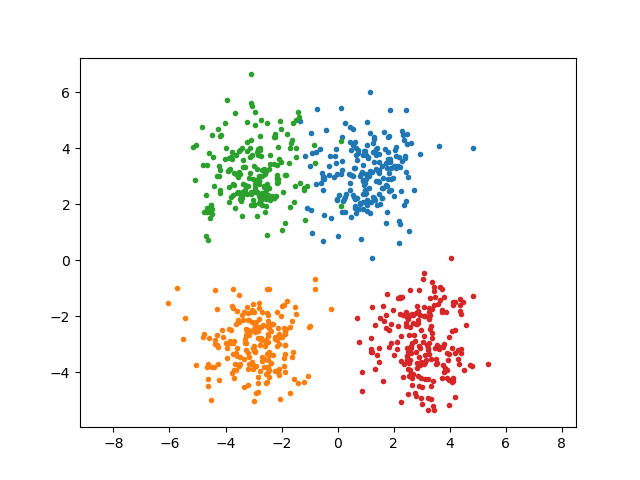}}
		\subcaptionbox{dist-2\label{fig:CI-dist2}}{\includegraphics[width=0.25\textwidth]{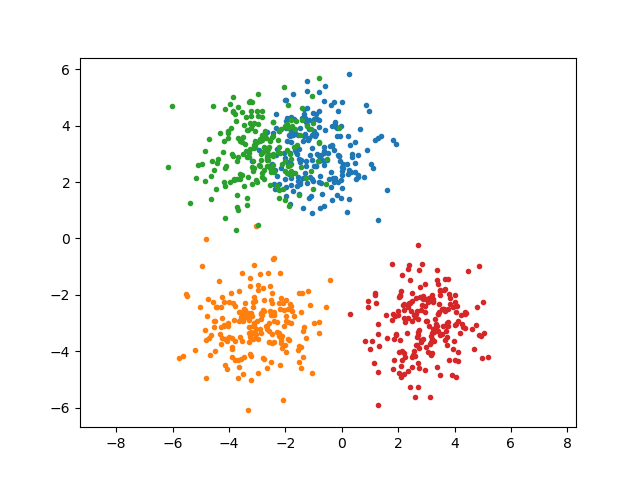}}
		\subcaptionbox{dist-1\label{fig:CI-dist1}}{\includegraphics[width=0.25\textwidth]{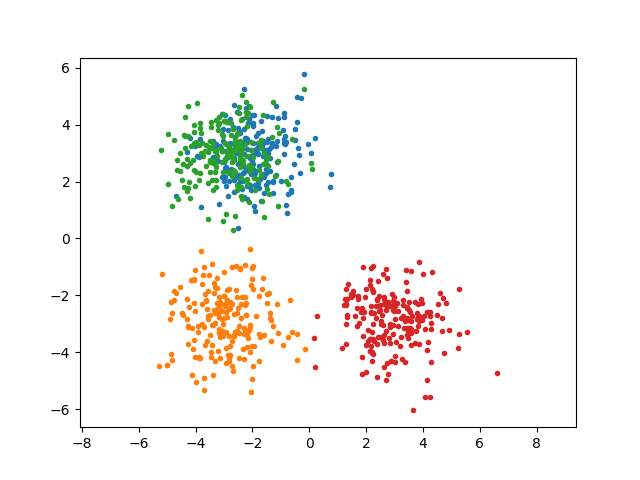}}
	\end{multicols}

	\caption{Cluster intermix based on distance between cluster centroids}
	\label{fig:CI_D}
\end{figure}

\begin{figure}[htb!]
	
	\centering
	\begin{multicols}{4}
		\subcaptionbox{\emph{cov-I}\label{fig:CI-spread1}}{\includegraphics[width=0.25\textwidth]{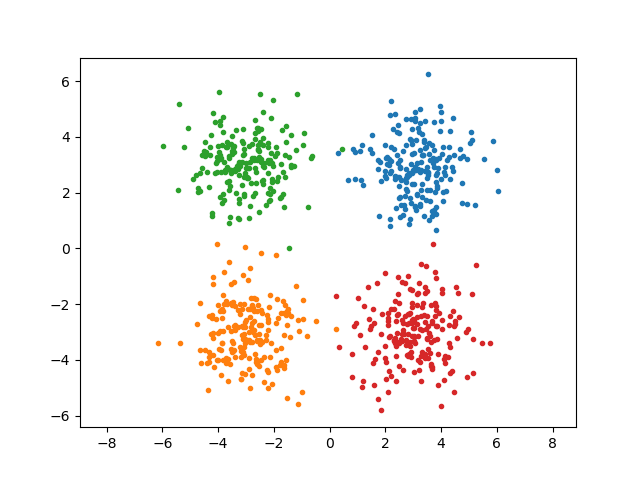}}
		\subcaptionbox{\emph{cov-5I}\label{fig:CI-spread5}}{\includegraphics[width=0.25\textwidth]{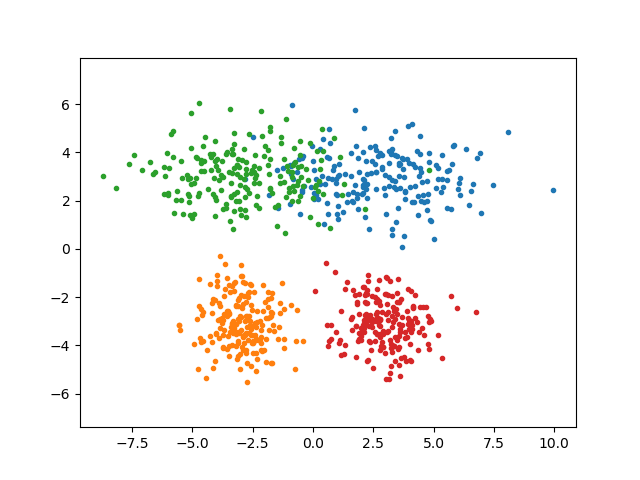}}
		\subcaptionbox{\emph{cov-10I}\label{fig:CI-spread10}}{\includegraphics[width=0.25\textwidth]{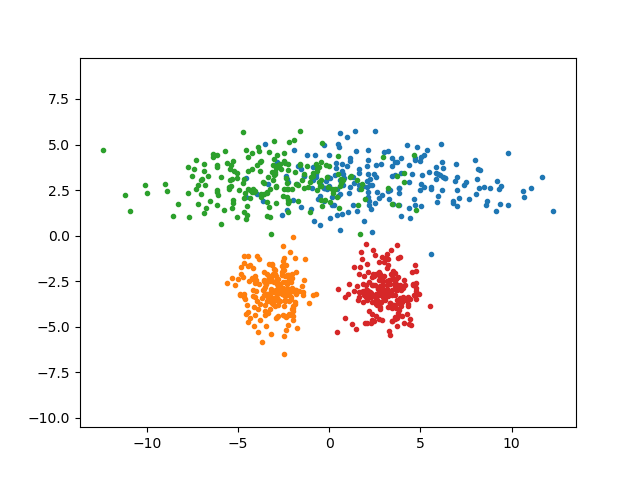}}
		\subcaptionbox{\emph{cov-15I}\label{fig:-spread15}}{\includegraphics[width=0.25\textwidth]{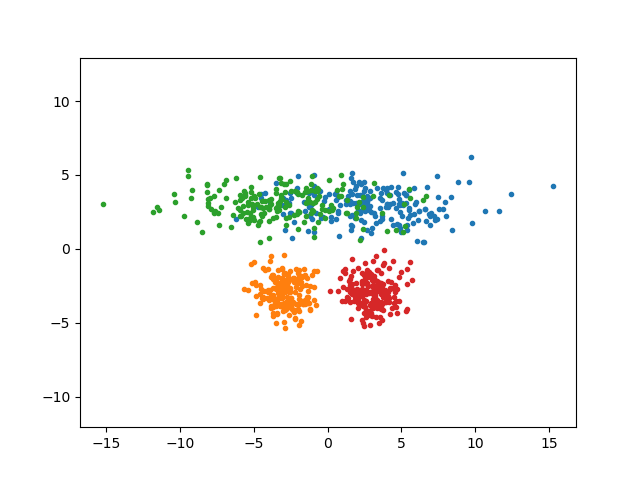}}
	\end{multicols}

	\caption{Cluster intermix based on $n*I$ covariance matrices of two clusters where n is an integer. Other two clusters have $I$ covariance matrices. $I$ represents Identity matrix.}
	\label{fig:CI_V}
\end{figure}

We consider Gaussian clusters and tune their parameters (centers, cluster shape, and spread, covariance matrices, etc.) to generate overlapping between two clusters. In this work, we tune ``distance between cluster centroids" and covariance matrices to create simulations. For this purpose, we generate four Gaussian clusters  which are spaced in a square with the length of its side 6. Each cluster has an identity covariance matrix. We introduce the intermix between clusters by moving one cluster centroid closer to another. Figure~\ref{fig:CI_D} presents the datasets with distances between two cluster centroids kept at  6, 4, 2 and 1. Additionally, we create another set of overlapping simulation based on changing covariance matrices. We multiply identity covariance matrices of two clusters by 5, 10 and 15, respectively. We have shown them in Figure~\ref{fig:CI_V}.
It is difficult to identify two clusters (shown in blue and green colors) separately in Figures~\ref{fig:CI-dist2} and~\ref{fig:CI-dist1} visually. Results (Table~\ref{tab:intermix}) corroborates with visual findings. It is noted that predictions of all the methods (except ``Hartigan" and $CV_{a}$)  synchronized with visual findings when overlapping is generated by changing the position of cluster centroids. However, every method has mixed responses 
in the second scenario. \emph{Jump}, and \emph{CNAK} correctly predict the cluster number as 4 in all the four cases. The \emph{CNAK} shows diverse responses in 50 trials against \emph{cov-10I} and \emph{cov-15I}.

\begin{table}[!htb]
	\centering
	\caption{Effects on performances for prediction of cluster number for cluster overlapping}
	\label{tab:intermix}
	\begin{tabular}{lcccccccc}
		\hline
		\multirow{2}{*}{Dataset} & \multicolumn{8}{c}{Comparison of seven approached in cluster number prediction} \\ \cline{2-9} 
		& \multicolumn{1}{c}{CH} & \multicolumn{1}{c}{Jump} & \multicolumn{1}{c}{Hartigan} & \multicolumn{1}{c}{Curvature} & \multicolumn{1}{c}{Silhouette} & \multicolumn{1}{c}{Gap} & \multicolumn{1}{c}{$CV_{a}$} & \multicolumn{1}{c}{CNAK} \\ \hline
		dist-6 & 4 & 4 & 42 & 4 & 4 & 4 & 4 & 4 \\
		dist-4 & 4 & 4 & 44 & 4 & 4 & 4 & 2 & 4, 3\\
		dist-2 & 3 & 3 & 36 & 3 & 3 & 4 & 2 & 3 \\
		dist-1 & 3 & 3 & 35 & 3 & 3 & 3 & 2 & 3\\ \hline 
		cov-I & 4 & 4 & 38 & 4 & 4 & 4 & 4 & 4 \\ 
		cov-5I & 5 & 4 & 48 & 17 & 4 & 4 & 4 & 4 \\ 
		cov-10I & 7 & 4 & 34 & 19 & 5 & 5 & 2 & \begin{tabular}[c]{@{}l@{}}4, 5, 2 \end{tabular} \\
		cov-15I & 6 & 4 & 32 & 4 & 5 & 6 & 2 & 4, 2 \\\hline
		average \emph{gs}& 0.726 & {\bf0.947} & 0.000 & 0.695  & 0.838 & 0.865 & 0.605 & 0.945 \\\hline
	\end{tabular}
\end{table}

Tables~\ref{tab:clustering_intermix_nh_dist} and \ref{tab:clustering_intermix_nh_cov} present the performance of clustering  with the predicted cluster number. In general, the accuracy of clustering with respect to the ground truth decreases with increasing overlapping. It is interesting to note that with increasing clustering accuracy from $k=3$ to $4$, the silhouette score decreases. This is an indicator of reduced ``between-cluster separation".

\begin{table}[htb!]
	\caption{Clustering results on simulation designed for cluster Intermix (non-hierarchical) by changing distance between cluster centroid}
	\label{tab:clustering_intermix_nh_dist}
	\makebox[\linewidth]{
		\begin{tabular}{llccccccccccc}
			\hline
			\multirow{2}{*}{Dataset} & \multirow{2}{*}{Methods} &K  & \multicolumn{2}{c}{ARI} & \multicolumn{2}{l}{NMI} & \multicolumn{2}{l}{Homogeneity} & \multicolumn{2}{l}{Completeness} & \multicolumn{2}{l}{Silhoutte} \\ \cline{4-13} 
			&  &  & \multicolumn{1}{c}{Mode} & \multicolumn{1}{c}{SM} & \multicolumn{1}{c}{Mode} & \multicolumn{1}{c}{SM} & \multicolumn{1}{c}{Mode} & \multicolumn{1}{c}{SM} & Mode & SM & \multicolumn{1}{c}{Mode} & \multicolumn{1}{c}{SM} \\ \hline
			\multirow{4}{*}{dist-6} & CNAK & 4 & {\bf 1.00 }&{\bf 1.00} &{\bf 1.00} &{\bf 1.00} &{\bf 1.00} &{\bf 1.00} &{\bf 1.00} &{\bf 1.00} & 0.70 & 0.70 \\
			& BA & 4 & 0.99 & 0.99 & 0.99 & 0.99 & 0.99 & 0.99 & 0.99 & 0.99 & 0.81 & 0.81 \\
			& RS & 4 & 0.99 & 0.99 & 0.99 & 0.99 & 0.99 & 0.99 & 0.99 & 0.99 & {\bf 0.81} &{\bf 0.81} \\
			& RPKM & 4 & 0.99 & 0.99 & 0.98  & 0.98 & 0.98  & 0.98 &0.98 &0.98 & 0.67 &0.66  \\ \hline

			\multirow{4}{*}{dist-4} & CNAK & 4 &{\bf 0.98} & {\bf 0.98} &{\bf 0.97} &{\bf 0.97} &{\bf 0.97} & {\bf 0.97} &{\bf 0.97} &{\bf 0.97} & 0.64 & 0.64 \\
			& BA & 4 & 0.96 & 0.93 & 0.94 & 0.92 & 0.94 & 0.92 & 0.94 & 0.92 & 0.76 & 0.74 \\
			& RS & 4 & 0.97 & 0.96 & 0.96 & 0.95 & 0.96 & 0.95 & 0.95 & 0.95 & {\bf 0.77} &{\bf 0.76} \\
			& RPKM & 4 & 0.90 & 0.93 &0.92  & 0.91 & 0.92 & 0.90 & 0.92 & 0.93 & 0.59 &0.59  \\ \hline

			\multirow{8}{*}{dist-2} & \multirow{2}{*}{CNAK} & 3 & 0.70 & 0.70 & 0.85 & 0.85 & 0.73 & 0.73 & 0.98 & 0.98 &{\bf 0.64} &{\bf 0.64} \\
			&  & 4 & {\bf 0.94} &{\bf 0.94} &{\bf 0.93} &{\bf 0.93} &{\bf 0.93} &{\bf 0.93} &{\bf 0.93} &{\bf 0.93} &{\bf 0.59} &{\bf 0.59} \\
			& \multirow{2}{*}{BA} & 3 & 0.70 & 0.69 & 0.85 & 0.83 & 0.73 & 0.72 & 0.98 & 0.96 & 0.51 & 0.49 \\
			&  & 4 & 0.69 & 0.72 & 0.79 & 0.78 & 0.77 & 0.76 & 0.83 & 0.81 & 0.43 & 0.48 \\
			& \multirow{2}{*}{RS} & 3 & {\bf 0.71} &{\bf 0.71 }&{\bf 0.85} &{\bf 0.85} &{\bf 0.74} &{\bf 0.74} &{\bf 0.98} &{\bf 0.98} & 0.51 & 0.51 \\
			&  & 4 & 0.62 & 0.62 & 0.78 & 0.78 & 0.73 & 0.73 & 0.83 & 0.83 & 0.36 & 0.36 \\
			& RPKM & 3 & 0.70 & 0.70 & 0.85 & 0.85 & 0.73 & 0.73 & 0.98 &0.98  &0.66  & 0.66 \\ 
			&  & 4 & 0.74 & 0.71 & 0.81 & 0.80 & 0.80 & 0.78 &0.83  & 0.84 & 0.52 & 0.52 \\\hline

			\multirow{8}{*}{dist-1} & \multirow{2}{*}{CNAK} & 3 & {\bf 0.71} & {\bf 0.71} &{\bf  0.86} &{\bf 0.86} &{\bf 0.74} &{\bf 0.74} &{\bf 0.99} & {\bf 0.99} &{\bf 0.66} &{\bf 0.66} \\
			&  & 4 & {\bf 0.99} & {\bf 0.99} & {\bf 0.99} &{\bf 0.99} &{\bf 0.99} & {\bf 0.99} &{\bf 0.99} &{\bf 0.99} &{\bf 0.54} &{\bf 0.54} \\
			& \multirow{2}{*}{BA} & 3 & 0.71 & 0.71 & 0.86 & 0.85 & 0.74 & 0.74 & 0.98 & 0.98 & 0.51 & 0.51 \\
			&  & 4 & 0.69 & 0.68 & 0.77 & 0.78 & 0.73 & 0.73 & 0.89 & 0.83 & 0.43 & 0.42 \\
			& \multirow{2}{*}{RS} & 3 & 0.71 & 0.71 & 0.86 & 0.86 & 0.74 & 0.74 & 0.99 & 0.99 & 0.51 & 0.50\\
			&  & 4 & 0.62 & 0.62 & 0.78 & 0.78 & 0.73 & 0.73 & 0.84 & 0.84 & 0.37 & 0.37 \\ 
			
			& \multirow{2}{*}{RPKM} & 3 & 0.71 &0.71  & 0.85 & 0.85 & 0.74 & 0.74 & 0.99 & 0.99 & 0.68 & 0.68 \\ 
			&  & 4 & 0.64 & 0.66 & 0.80 &0.81  &0.74  & 0.74 & 0.86 &0.88  &0.59  & 0.59 \\\hline 
		\end{tabular}
	}
\end{table}

\begin{table}[htb!]
	\caption{Clustering results on simulation designed for cluster Intermix (non-hierarchical) by changing covariance matrix}
	\label{tab:clustering_intermix_nh_cov}
	\makebox[\linewidth]{
		\begin{tabular}{llccccccccccc}
			\hline
			\multirow{2}{*}{Dataset} & \multirow{2}{*}{Methods} &  & \multicolumn{2}{c}{ARI} & \multicolumn{2}{l}{NMI} & \multicolumn{2}{l}{Homogeneity} & \multicolumn{2}{l}{Completeness} & \multicolumn{2}{l}{Silhoutte} \\ \cline{4-13} 
			&  &  & \multicolumn{1}{c}{Mode} & \multicolumn{1}{c}{SM} & \multicolumn{1}{c}{Mode} & \multicolumn{1}{c}{SM} & \multicolumn{1}{c}{Mode} & \multicolumn{1}{c}{SM} & Mode & SM & \multicolumn{1}{c}{Mode} & \multicolumn{1}{c}{SM} \\ \hline
			
			\multirow{4}{*}{cov\_I} & CNAK & 4 & {\bf 0.99} & {\bf 0.99} &{\bf 0.99} &{\bf 0.99} &{\bf 0.99} &{\bf 0.99} &{\bf 0.99} &{\bf 0.99} & 0.69 & 0.69 \\
			& BA & 4 & 0.98 & 0.98 & 0.97 & 0.97 & 0.97 & 0.97 & 0.97 & 0.97 & 0.80 & 0.79 \\
			& RS & 4 & 0.98 & 0.98 & 0.97 & 0.97 & 0.97& 0.97 & 0.97 & 0.97 &{\bf 0.80} &{\bf 0.80} \\
			& RPKM & 4 &0.99  & 0.95 & 0.98 &0.95  &0.98  &0.94  & 0.98 & 0.96 & 0.66 &0.64  \\ \hline
			
			\multirow{4}{*}{cov\_5I} & CNAK & 4 & 0.91 & 0.91 & 0.90 & 0.90 & 0.906 & 0.905 & 0.90 & 0.90 & 0.59 & 0.59 \\
			& BA & 4 & {\bf 0.98} & {\bf 0.98} & {\bf 0.97} & {\bf 0.97} & {\bf 0.97} & {\bf 0.97} & {\bf 0.97} & {\bf 0.97} & {\bf 0.80} & {\bf 0.80} \\
			& RS & 4 & 0.87 & 0.88 & 0.88 & 0.88 & 0.88 & 0.88 & 0.88 & 0.88 & 0.69 & 0.69 \\
			& RPKM & 4 & 0.85 &0.74  & 0.86 &0.79  & 0.85 & 0.77 & 0.86 & 0.82 & 0.57 & 0.50 \\ \hline

			\multirow{4}{*}{cov\_10I} & CNAK & 4 &{\bf 0.81} & {\bf 0.81} &{\bf 0.84} &{\bf 0.83} &{\bf 0.83} &{\bf 0.83} &{\bf 0.84} &{\bf 0.84} & 0.55 & 0.55 \\
			& BA & 4 & 0.78 & 0.68 & 0.78 & 0.72 & 0.79 & 0.72 & 0.77 & 0.73 & 0.58 & 0.47 \\
			& RS & 4 & 0.78 & 0.78 & 0.81 & 0.81 & 0.81 & 0.81 & 0.81 & 0.81 & {\bf 0.59} &{\bf 0.59} \\
			& RPKM & 4 & 0.46 & 0.49 & 0.63 & 0.64 & 0.52 &0.55  &0.76  & 0.74 & 0.34 & 0.37 \\ \hline
			
			\multirow{4}{*}{cov\_15I} & CNAK & 4 & {\bf 0.79} & {\bf 0.79} & {\bf 0.82} &{\bf 0.82} &{\bf 0.82} &{\bf 0.82} &{\bf 0.82 }&{\bf 0.82} & 0.56 & 0.56 \\
			& BA & 4 & 0.73 & 0.64 & 0.77 & 0.69 & 0.77 & 0.69 & 0.77 & 0.70 & 0.54 & 0.43 \\
			& RS & 4 & 0.77 & 0.77 & 0.81 & 0.81 & 0.81 & 0.81 & 0.81 & 0.81 & {\bf 0.59} & {\bf 0.58} \\
			& RPKM & 4 & 0.47 & 0.50 & 0.63 & 0.62 & 0.51 &0.55  &0.77  &0.71  &0.35  &0.39  \\ \hline
		\end{tabular}
	}
\end{table}

\subsubsection{Study on high dimensional dataset (sim-g2)}
\begin{table}[htb!]
	\centering
	\caption{A brief description of synthetic high  dimensional dataset. ($\lambda_{max}$, $\tau$, $c$) are the parameters for the determination of the size of sampled dataset as discussed in section~\ref{sec:sample_size}.} 
	\begin{tabular}{lccccccc}
		\hline
		Dataset & Instances & Features & Classes & $\lambda_{max}$    & $\tau$ & c           & \begin{tabular}[c]{@{}c@{}}Sample size\\  (\%)\end{tabular} \\ \hline	dim32   & 1024 & 32   & 16 & 17866.16  & 4 & 11.56 & 34  \\ 
		dim64   & 1024 & 64   & 16 & 22870.62  & 4 & 12.30 & 36 \\ 
		dim128  & 1024 & 128  & 16 & 34119.84  & 4 & 13.59 & 41   \\ 
		dim256  & 1024 & 256  & 16 & 67076.31  & 3 & 16.09 & 13 \\
		dim512  & 1024 & 512  & 16 & 122947.96 & 3 & 49.72 & 16     \\ 
		dim1024 & 1024 & 1024 & 16 & 235344.67 & 3 & 61.74 & 19   \\ \hline
	\end{tabular}
	\label{tab:dims}
\end{table}

We consider six high dimensional datasets~\cite{DIMsets} dim32 to dim1024\footnote{For details please visit \url{http://cs.joensuu.fi/sipu/datasets/}} with higher dimensionality varying from 32 to 1,024. Clusters are distinctly separated. We summarize its features in Table~\ref{tab:dims}.
We examine the robustness of different methods against the large dimensional dataset. We find in Table~\ref{tab:dim_CN} that \emph{CH, Silhouette,} and \emph{CNAK} are capable of predicting true cluster numbers whereas others fail. However, the results of clustering in Table~\ref{tab:clustering_dimension} shows that  \emph{CNAK} performs comparatively better than other methods.
Additionally, we add white Gaussian noise such that {\emph SNR}\footnote{$SNR=\frac{\norm{S}}{\norm{N}}$ where $S$ and $N$ represent data and noise respectively. \norm{.} denotes \emph{Frobenius} norm.} value ranges from 100dB to 30dB. There is no such change in the results while perturbing this simulation with noise.

\begin{table}[htb]
	\centering
	\caption{Effects on Performances for prediction of cluster number with increasing feature dimension. Overflow encountered in double\_scalars for jump dataset for dimension of feature greater than 64.}
	\label{tab:dim_CN}
	\begin{tabular}{lccccccccc}
		\hline
		\multirow{2}{*}{Dataset} & \multirow{2}{*}{K} & \multicolumn{8}{l}{Comparison of seven approaches in Cluster Number Prediction} \\ \cline{3-10} 
		&  & \emph{CH} & \emph{Jump}& \emph{Hartigan} & \emph{Curvature} & \emph{Silhouette} & \emph{Gap} & $CV_{a}$ & \emph{CNAK} \\ \hline
		dim32 & 16 & 16 & 14& 16& 14 & 16 & 23 & 11 & 16 \\ 
		dim64 & 16 & 16 & 14&17 & 14 & 16 & 24 & 13 & 16 \\ 
		dim128 & 16 & 16 & - & 18 & 14 & 16 & 22 & 21 & 16 \\ 
		dim256 & 16 & 16 & - & 16 & 14 & 16 & 25 & 21 & 16 \\ 
		dim512 & 16 & 16 & - & 16 &14 & 16 & 24 & 21 & 16 \\ 
		dim1024 & 16 & 16 & - & 16&14 & 16 & 23 & 23 & 16 \\ \hline
		\multicolumn{2}{l}{Average \emph{gs} score}&{\bf 1.00}&-&0.857 &0.37 & {\bf 1.00} & 0.00 & 0.01 & {\bf 1.00}\\\hline
	\end{tabular}
\end{table}

\begin{table}[htb!]
	\centering
	\caption{Clustering results on the simulation designed for checking scalability (increasing number of dimensions). Results are shown for two k appear in sorted score (ascending order)}
	
	\makebox[\linewidth]{
		\begin{tabular}{llcccccccccc}
			\hline
			\multirow{2}{*}{Dataset} & \multirow{2}{*}{Methods} & \multicolumn{2}{c}{ARI} & \multicolumn{2}{c}{NMI} & \multicolumn{2}{c}{Homogeneity} & \multicolumn{2}{c}{Completeness} & \multicolumn{2}{c}{Silhoutte}\\ \cline{3-12} 
			&  & Mode & SM & Mode & SM & Mode & SM & Mode & SM & Mode & SM \\ \hline
			\multirow{4}{*}{dim32} & CNAK & {\bf 1.00} &{\bf 1.00} &{\bf 1.00} & {\bf 1.00} & {\bf 1.00} & {\bf 1.00} &{\bf  1.00} &{\bf 1.00} & {\bf 0.99}&{\bf 0.99} \\ 
			& BA & 0.94 & 0.91 & 0.98 & 0.98 & 0.97 & 0.96 & 1.00 & 0.99&0.87&0.84 \\ 
			& RS & 0.74 & 0.71 & 0.93 & 0.92 & 0.87 & 0.85 & 1.00 & 1.00&0.66&0.68 \\ 
			& RPKM & 0.17 & 0.18 & 0.58 & 0.58 & 0.35 & 0.35 & 1.00 & 0.98&0.21&0.20 \\ \hline
			\multirow{4}{*}{dim64} & {\bf CNAK} & {\bf 1.00} &{\bf 1.00} &{\bf 1.00} &{\bf 1.00} &{\bf 1.00} &{\bf 1.00} &{\bf 1.00} &{\bf 1.00}& {\bf 0.99}& {\bf 0.99} \\ 
			& BA & 0.88 & 0.90 & 0.97 & 0.97 & 0.97 & 0.95 & 1.00 & 0.99&0.89&0.85 \\ 
			& RS & 0.32 & 0.32 & 0.76 & 0.75 & 0.58 & 0.57 & 1.00 & 1.00&0.41&0.38 \\ 
			& RPKM & 0.11 & 0.13 & 0.55 & 0.53 & 0.30 & 0.29 & 1.00 & 0.97&0.19&0.17 \\ \hline
			\multirow{4}{*}{dim128} & {\bf CNAK} & {\bf 1.00} & {\bf 1.00}& {\bf 1.00} &{\bf 1.00} &{\bf 1.00} &{\bf 1.00} &{\bf 1.00} &{\bf 1.00}& {\bf 0.99}& {\bf 0.99} \\ 
			& BA & 0.94 & 0.92 & 0.98 & 0.98 & 0.97 & 0.96 & 1.00 & 0.99&0.91&0.88 \\ 
			& RS & 0.53 & 0.60 & 0.86 & 0.86 & 0.74 & 0.74 & 1.00 & 0.99&0.58&0.55 \\ 
			& RPKM & 0.11 & 0.13 & 0.54 & 0.53 & 0.30 & 0.29 & 1.00 & 0.98&0.18&0.16 \\ \hline
			\multirow{4}{*}{dim256} & {\bf CNAK} & {\bf 1.00} & {\bf 1.00} & {\bf 1.00} & {\bf 1.00} & {\bf 1.00} & {\bf 1.00} & {\bf 1.00} & {\bf 1.00} & {\bf 0.99} &{\bf 0.99} \\ 
			& BA & 0.88 & 0.92 & 0.97 & 0.98 & 0.94 & 0.96 & 1.00 & 0.99&0.92&0.89 \\ 
			& RS & 0.60 & 0.57 & 0.89 & 0.87 & 0.79 & 0.76 & 1.00 & 1.00&0.59&0.57 \\ 
			& RPKM & 0.11 & 0.14 & 0.56 & 0.55 & 0.31 & 0.31 & 1.00 & 0.99&0.18&0.17 \\ \hline
			\multirow{4}{*}{dim512} & {\bf CNAK} & {\bf 1.00} & {\bf 1.00} & {\bf 1.00} & {\bf 1.00} & {\bf 1.00} & {\bf 1.00} & {\bf 1.00} & {\bf 1.00} & {\bf 0.99} & {\bf 0.99} \\ 
			& BA & 0.88 & 0.90 & 0.97 & 0.97 & 0.94 & 0.95 & 1.00 & 0.99&0.92&0.88 \\ 
			& RS & 0.37 & 0.35 & 0.77 & 0.76 & 0.54 & 0.59 & 1.00 & 0.99&0.47&0.38 \\ 
			& RPKM & 0.11 & 0.14 & 0.58 & 0.53 & 0.29 & 0.30 & 1.00 & 0.99&0.18&0.17 \\ \hline
			\multirow{4}{*}{dim1024} & {\bf CNAK} & {\bf 1.00} & {\bf 1.00} & {\bf 1.00} & {\bf 1.00} & {\bf 1.00} & {\bf 1.00} & {\bf 1.00} & {\bf 1.00} & {\bf 0.99} & {\bf 0.99} \\ 
			& BA & 0.94 & 0.90 & 0.98 & 0.97 & 0.97 & 0.95 & 1.00 & 0.99&0.93&0.90 \\ 
			& RS & 0.63 & 0.58 & 0.88 & 0.87 & 0.79 & 0.76 & 1.00 & 1.00&0.61&0.58 \\ 
			& RPKM & 0.17 & 0.16 & 0.60 & 0.57 & 0.36 & 0.33 & 1.00 & 0.99&0.18&0.19 \\ \hline
		\end{tabular}
	}
	\label{tab:clustering_dimension}
\end{table}

\subsubsection{Study on Shape clusters (sim-g3)}

\begin{figure}
	\begin{multicols}{4}
		\subcaptionbox{Aggregation\label{fig:Aggregation}}{\includegraphics[width=0.25\textwidth]{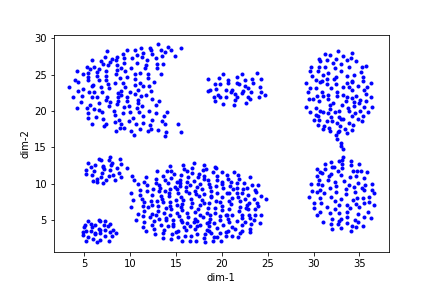}}
		\subcaptionbox{Compound\label{fig:Compound}}{\includegraphics[width=0.25\textwidth]{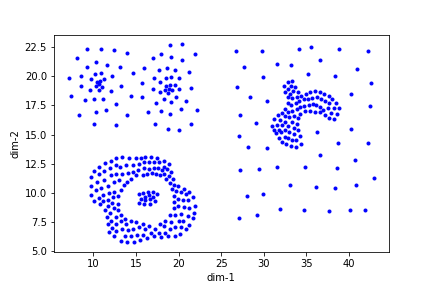}}
		\subcaptionbox{Pathbased\label{fig:pathbased}}{\includegraphics[width=0.25\textwidth]{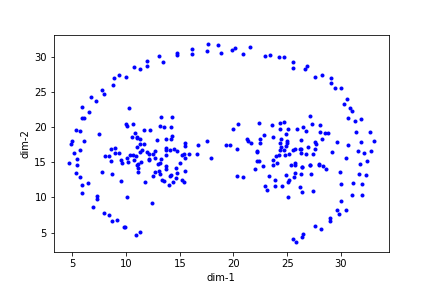}}
		\subcaptionbox{Spiral\label{fig:spiral}}{\includegraphics[width=0.25\textwidth]{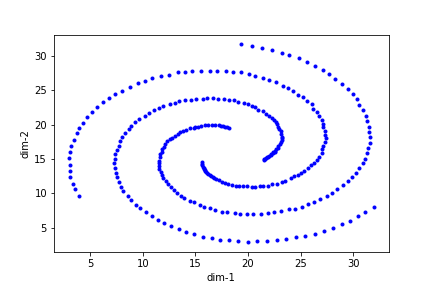}}
	\end{multicols}	
	
	\begin{multicols}{4}
		\subcaptionbox{Flame\label{fig:flame}}{\includegraphics[width=0.25\textwidth]{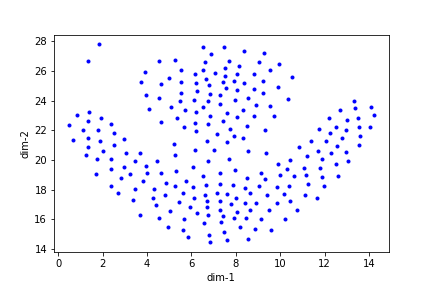}}
		\subcaptionbox{Jain\label{fig:jain}}{\includegraphics[width=0.25\textwidth]{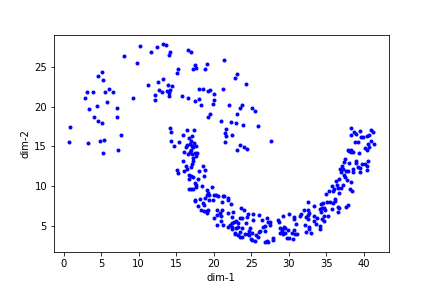}}
		\subcaptionbox{R15\label{fig:R15}}{\includegraphics[width=0.25\textwidth]{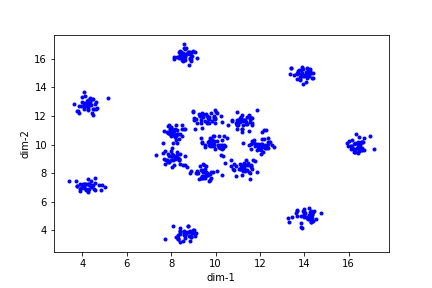}}
		\subcaptionbox{D31\label{fig:D31}}{\includegraphics[width=0.25\textwidth]{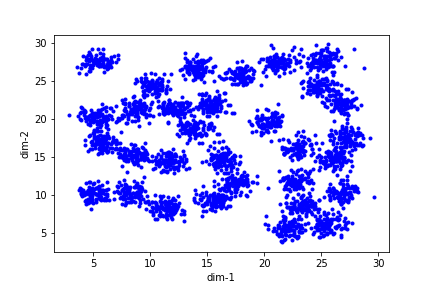}}
	\end{multicols}			

	\caption{Datasets of different shape structure}
	\label{fig:shape}
\end{figure}

\begin{table}[htb!]
	\centering
	\caption{A brief description of synthetic shape dataset. ($\lambda_{max}$, $\tau$, $c$) are the parameters for the determination of the size of sampled dataset as discussed in section~\ref{sec:sample_size}.} 
	\begin{tabular}{lccccccc}
		\hline
		Dataset     & Instances & Features & Classes & $\lambda_{max}$    & $\tau$ & c    & \begin{tabular}[c]{@{}c@{}}Sample size\\ (\%)\end{tabular} \\ \hline
		Aggregation~\cite{Aggregation}& 788 & 2 & 7  & 98.49  & 16  & 1.33 & 21 \\ 
		Compound~\cite{Compound}   & 399 & 2 & 6  & 97.68  & 16  & 1.33 & 35 \\ 
		Path-based~\cite{pathbased_spiral}  & 300 & 2 & 3  & 68.12  & 16  & 1.30 & 34  \\ 
		Spiral~\cite{pathbased_spiral}      & 312 & 2 & 3  & 53.29  & 16  & 1.28 & 29  \\ 
		Flame~\cite{Flame}       & 240 & 2 & 2  & 11.47  & -   & 0.2  & 82   \\ 
		Jain~\cite{Jain}       & 373 & 2 & 2  & 103    & 16  & 1.34 & 39  \\ 
		R15~\cite{D31_R15}       & 600 & 2 & 15 & 10.67  & -   & 0.2  & 63  \\ 
		D31~\cite{D31_R15}         & 3100 & 2 & 31 & 55.75  & -  & 0.2 & 63 \\ \hline
	\end{tabular}
	\label{tab:shape}
\end{table}

This simulation is used to compare the methods on different shape  datasets~\cite{Aggregation,Compound,pathbased_spiral,D31_R15,Flame,Jain}. We provide their brief description in Table~\ref{tab:shape}.
Table~\ref{tab:CNshape}  affirms the limitation of \emph{CNAK} towards different shape identification, which is an inherent limitation of the \emph{k-means} algorithm. \emph{R15} and \emph{D31} may contain many small spherical clusters as shown in Figure~\ref{fig:R15} and \ref{fig:D31}, respectively. This might be the reason behind successful prediction for these two datasets by \emph{CNAK}, \emph{CH}, and \emph{silhouette}.  Clustering outcome in Table~\ref{tab:clustering_shape} confirms the findings with $ARI$ and $NMI$ $>0.9$ in \emph{D31} and \emph{R15}.

\begin{table}[htb!]
	\centering
	\caption{Performances of cluster number prediction in different shape datasets}
	\begin{tabular}{lccccccccc}
		\hline
		\multirow{2}{*}{Dataset} & \multirow{2}{*}{K} & \multicolumn{8}{l}{Comparison of seven approaches in Cluster Number Prediction} \\ \cline{3-10} 
		&  & CH & Jump &Hartigan& Curvature & Silhouette & Gap & $CV_{a}$ & CNAK \\ \hline
		Aggregation~\cite{Aggregation} & 7 & 26 & 2&37 & 30 & 4 & 16 & 2 & 6 \\ 
		Compound~\cite{Compound} & 6 & 2 & 3& 26 & 10 & 2 & 11 & 2 & 3 \\ 
		pathbased~\cite{pathbased_spiral} & 3 & 26 & 9&28 & 28 & 3 & 2 & 2 & 2 \\ 
		spiral~\cite{pathbased_spiral} & 3 & 26 & 26&18 & 19 & 25 & 1 & 20 & 8 \\ 
		Flame~\cite{Flame} & 2 & 8 & 2& 15 & 24 & 4 & 8 & 11 & 7 \\ 
		Jain~\cite{Jain} & 2 & 20 & 4& 21 & 25 & 7 & 13 & 2 & 2 \\ 
		R15~\cite{D31_R15} & 15 & 15 & 13& 20 & 13 & 15 & 18 & 7 & 15 \\ 
		D31~\cite{D31_R15} & 31 & 31 & 25& 57 & 29 & 31 & 36 & 2 & 31 \\ \hline
		\multicolumn{2}{l}{Average \emph{gs} score}&0.25& 0.23&0.00 & 0.09 & 0.43 & 0.16 & 0.14 & {\bf 0.58}\\\hline
	\end{tabular}
	\label{tab:CNshape}
\end{table}

\begin{table}[htb!]
	\caption{Clustering results on shape dataset with cluster number predicted by \emph{CNAK} and ground truth}
	\makebox[\linewidth]{
		\begin{tabular}{llccccccccccc}
			\hline
			\multirow{2}{*}{Dataset} & \multirow{2}{*}{Methods} & \multicolumn{1}{l}{\multirow{2}{*}{K}} & \multicolumn{2}{c}{ARI} & \multicolumn{2}{c}{NMI} & \multicolumn{2}{c}{Homogeneity} & \multicolumn{2}{c}{Completeness} & \multicolumn{2}{c}{Silhoutte} \\ \cline{4-13} 
			&  & \multicolumn{1}{l}{} & \multicolumn{1}{l}{Mode} & \multicolumn{1}{l}{SM} & \multicolumn{1}{l}{Mode} & \multicolumn{1}{l}{SM} & \multicolumn{1}{l}{Mode} & \multicolumn{1}{l}{SM} & \multicolumn{1}{l}{Mode} & \multicolumn{1}{l}{SM} & \multicolumn{1}{l}{Mode} & \multicolumn{1}{l}{SM} \\ \hline
			\multirow{8}{*}{Aggregation} & \multirow{2}{*}{CNAK} & 6 & {\bf 0.80} &{\bf 0.80} & 0.87 & 0.87 & {\bf 0.88} & {\bf 0.88} & {\bf 0.87} & {\bf 0.87} & {\bf 0.48} & \multicolumn{1}{l}{\bf 0.48} \\ 
			&  & 7 & 0.72 & 0.72 & 0.83 & 0.83 & 0.88 & 0.88 & 0.79 & 0.79 & 0.47 & \multicolumn{1}{l}{0.47} \\ 
			& \multirow{2}{*}{BA} & 6 & 0.62 & 0.61 & 0.74 & 0.75 & 0.74 & 0.75 & 0.74 & 0.75 & 0.34 & 0.37 \\ 
			&  & 7 & 0.67 & 0.65 & 0.81 & 0.80 & 0.86 & 0.85 & 0.77 & 0.76 & 0.42 & 0.41 \\ 
			& \multirow{2}{*}{RS} & 6 & 0.79 & 0.79 & {\bf 0.88} & {\bf 0.88} & {\bf 0.88} & {\bf 0.88} & {\bf 0.87} & {\bf 0.87} & {\bf 0.48} & {\bf 0.48} \\ 
			&  & 7 & 0.73 & 0.73 & 0.84 & 0.84 & 0.88 & 0.88 & 0.79 & 0.79 & 0.47 & 0.47 \\ 
			& \multirow{2}{*}{RPKM} & 6 & 0.43 & 0.41 & 0.61 & 0.61 & 0.49 & 0.47 & 0.77 & 0.77 & 0.44 & 0.43 \\ 
			&  & 7 & 0.43 & 0.40 & 0.61 & 0.60 & 0.49 & 0.47 & 0.77 & 0.77 & 0.44 & 0.43 \\ \hline
			\multirow{8}{*}{Compound} & CNAK & 3 & {\bf 0.73} & {\bf 0.73} & {\bf 0.79} & {\bf 0.79} & {\bf 0.65} & {\bf 0.65} & {\bf 0.96} & {\bf 0.96} & {\bf 0.61} & {\bf 0.61} \\ 
			&  & 6 & 0.55 & 0.55 & 0.70 & 0.69 & 0.72 & 0.73 & 0.67 & 0.66 & 0.41 & 0.41 \\ 
			& \multirow{2}{*}{BA} & 3 & 0.73 & 0.73 & 0.79 & 0.79 & 0.65 & 0.65 & 0.96 & 0.96 & 0.61 & 0.61 \\ 
			&  & 6 & 0.56 & 0.56 & 0.70 & 0.70 & 0.71 & 0.71 & 0.68 & 0.69 & 0.38 & 0.38 \\ 
			& \multirow{2}{*}{\bf RS} & 3 & {\bf 0.73} & {\bf 0.73} & {\bf 0.79} & {\bf 0.79} & {\bf 0.65} & {\bf 0.65} & {\bf 0.96} & {\bf 0.96} & {\bf 0.61 } & {\bf 0.61} \\ 
			&  & 6 & {\bf 0.59} & {\bf 0.55} & {\bf 0.74} & {\bf 0.71} & {\bf 0.77} &{\bf  0.74} & {\bf 0.71} & {\bf 0.68} & {\bf 0.43} & {\bf 0.42} \\ 
			& \multirow{2}{*}{RPKM} & 3 & 0.74 & 0.65 & 0.82 & 0.76 & 0.67 & 0.60 & 1.00 & 0.97 & 0.57 & 0.57 \\ 
			&  & 6 & 0.74 & 0.58 & 0.82 & 0.72 & 0.67 & 0.54 & 1.00 & 0.96 & 0.57 & 0.58 \\ \hline
			\multirow{8}{*}{Pathbased} & \multirow{2}{*}{CNAK} & 2 & {\bf 0.40} & {\bf 0.40} & {\bf 0.50} & {\bf 0.50} & {\bf 0.40} & {\bf 0.40} & {\bf 0.63} & {\bf 0.63} & {\bf 0.51} & {\bf 0.51} \\ 
			&  & 3 & {\bf 0.46} &{\bf 0.46} & {\bf 0.55} & {\bf 0.55} & {\bf 0.52} & {\bf 0.52} & {\bf 0.58} & {\bf 0.58} & {\bf 0.54} & {\bf 0.54} \\ 
			& \multirow{2}{*}{BA} & 2 & {\bf 0.40} & {\bf 0.40} & {\bf 0.50} & {\bf 0.50} & {\bf 0.40} & {\bf 0.40} & {\bf 0.63} & {\bf 0.63} & {\bf 0.51} & {\bf 0.51} \\ 
			&  & 3 & {\bf 0.46} & {\bf 0.46} & {\bf 0.54} & {\bf 0.54} & {\bf 0.52} & {\bf 0.51} & {\bf 0.58} & {\bf 0.57} & {\bf 0.53} & {\bf 0.53} \\ 
			& \multirow{2}{*}{RS} & 2 & {\bf 0.40} & {\bf 0.40} & {\bf 0.50} & {\bf 0.50} &{\bf 0.40} & {\bf 0.40} & {\bf 0.63} & {\bf 0.63} & {\bf 0.51} & {\bf 0.51} \\ 
			&  & 3 & {\bf 0.46} & {\bf 0.46} & {\bf 0.55} & {\bf 0.55} & {\bf 0.52} & {\bf 0.52} & {\bf 0.58} & {\bf 0.58} & {\bf 0.54} & {\bf 0.54} \\ 
			& \multirow{2}{*}{RPKM} & 2 & 0.39 & 0.37 & 0.50 & 0.46 & 0.43 & 0.41 & 0.63 & 0.52 & 0.51 & 0.45 \\ 
			&  & 3 & 0.37 & 0.36 & 0.44 & 0.44 & 0.43 & 0.41 & 0.46 & 0.47 & 0.43 & 0.42 \\ \hline
			\multirow{4}{*}{Jain} & {\bf CNAK} & 2 & {\bf 0.55} & {\bf 0.55} & {\bf 0.51} &{\bf  0.51} & {\bf 0.55} & {\bf 0.55} & {\bf 0.47} & {\bf 0.47} &  {\bf 0.50 }& {\bf 0.50 }\\ 
			& BA & 2 & 0.54 & 0.54 & 0.50 & 0.50 &  0.54 & 0.54 & {\bf 0.47} & {\bf 0.47} & {\bf 0.50} & {\bf 0.50} \\ 
			& RS & 2 & 0.55 & 0.55 & 0.51 & 0.51 & 0.55 & 0.55 & 0.47 & 0.47 & 0.50 & 0.50 \\ 
			& RPKM & 2 & 0.32 & 0.32 & 0.34 & 0.34 & 0.44 & 0.44 & 0.26 & 0.26 & 0.32 & 0.32 \\ \hline
			\multirow{4}{*}{R15} &{\bf CNAK} & 15 &{\bf 0.99} &{\bf 0.99} &{\bf 0.99} & {\bf 0.99 }& {\bf 0.99} & {\bf 0.99 }& {\bf 0.99} & {\bf 0.99} & {\bf 0.75} & {\bf 0.75} \\ 
			& BA & 15 & 0.86 & 0.83 & 0.92 & 0.91 & 0.98 & 0.96 & 0.87 & 0.86 & 0.48 & 0.44 \\ 
			& RS & 15 & 0.91 & 0.88 & 0.96 & 0.95 & 0.95 & 0.93 & 0.98 & 0.97 & 0.71 & 0.69 \\ 
			& RPKM & 15 & 0.18 & 0.17 & 0.53 & 0.53 & 0.33 & 0.33 & 0.85 & 0.86 & 0.27 & 0.26 \\ \hline
			\multirow{4}{*}{D31} &{\bf CNAK }& 31 & {\bf 0.95} & {\bf0.95} & {\bf 0.96} & {\bf0.96} & {\bf 0.97} & {\bf 0.97} & {\bf 0.97} & {\bf 0.97} & {\bf 0.58} & {\bf 0.58} \\ 
			& BA & 31 & 0.52 & 0.51 & 0.82 & 0.81 & 0.77 & 0.76 & 0.86 & 0.86 & 0.28 & 0.27 \\ 
			& RS & 31 & 0.72 & 0.69 & 0.90 & 0.88 & 0.87 & 0.85 & 0.92 & 0.92 & 0.49 & 0.48 \\ 
			& RPKM & 31 & 0.09 & 0.09 & 0.50 & 0.50 & 0.27 & 0.27 & 0.90 & 0.90 & 0.39 & 0.38 \\ \hline
		\end{tabular}
	}
	\label{tab:clustering_shape}
\end{table}

\subsection{Real-World Datasets}
To check the performance of the \emph{CNAK} algorithm in real-world datasets, we have selected \emph{iris, wine, seed, breast-cancer, landsat-satellite,} and \emph{magic} from the \emph{UCI} repository~\citep{Dua:2017}. We provide a brief description of these datasets in Table~\ref{tab:DD}.
\begin{table}[htb!]
	
	\caption{A brief description of real-world datasets. ($\lambda_{max}$, $\tau$, $c$) are the parameters for the determination of the size of sampled dataset. 
	} 	
	\begin{tabular}{lcccccc}
		\hline
		Dataset & Instances & Feature Dimension & $\lambda_{max}$&$\tau$ & $c$& \begin{tabular}[c]{@{}c@{}}Sample Size\\     (\%)\end{tabular} \\ \hline
		\emph{iris} & 150 & 4 &4.22 &- & 0.6& 24 \\
		\emph{wine} & 400 & 13&99201.79 &3 &46.29 & 49 \\
		{seed} & 480 & 7 &10.79 &- &0.6 & 35 \\
		\emph{breast-cancer} & 720 & 9& 49.047& 16&1.275 & 14 \\
		\emph{landsat satellite}&6435&36&5757.44&16&1.72&54\\ 
		\emph{magic} & 19020 & 10 &6579.79 &8 &3 &12.8 \\ \hline

	\end{tabular}
	\label{tab:DD}
\end{table}

\subsubsection{Prediction of cluster number}
To check the stability of \emph{CNAK}, we have executed it 50 times separately on the same dataset, and observe its performance in each of the $50$ trials. We have enlisted our findings in Table \ref{tab:CN}. \emph{Seed}, \emph{wine}, \emph{breast-cancer,} and \emph{magic} datasets have produced cluster numbers similar to their associated class labels by \emph{CNAK}. We observe that \emph{CNAK} predicts $3$ as cluster number in \emph{iris} dataset for 15 times. We emphasize on majority voting by 50 different execution for selecting absolute cluster number. Although, we may not ignore this estimated value as  $3$ is estimated in 30\% of total trials. However, \emph{landsat} \emph{satellite} predicts two as the cluster number instead of six. According to the dataset, it contains red soil, cotton crop, grey soil, damp grey soil,  soil with vegetation stubble, and very damp grey soil classes. This indicates the two major classes: soil and crop. Overall, \emph{CH} and \emph{CNAK} perform better than other techniques in estimating  cluster number for small-scale real-world datasets.\\

\begin{table*}[htb!]		
	\centering
	\caption{Cluster Number Prediction on Real-world datasets. \emph{ls} represents landset-satellite, and \emph{bc} represents breast-cancer dataset}.
	\label{tab:CN}
	\makebox[\linewidth]{
		\begin{tabular}{lccccccccc}
			\hline
			\multirow{2}{*}{Dataset} & \multirow{2}{*}{\begin{tabular}[c]{@{}c@{}}True\\$k$\end{tabular}} & \multicolumn{8}{c}{Comparison of seven approaches in Cluster Number Prediction} \\ \cline{3-10} 
			&  & CH & \emph{Jump} & Hartigan &\begin{tabular}[c]{@{}l@{}} Curvature \end{tabular}& \begin{tabular}[c]{@{}l@{}} Silhouette\end{tabular} & Gap& \begin{tabular}[c]{@{}l@{}} $CV_{a}$\end{tabular} & CNAK \\ \hline
			\emph{iris} & 3 & 3 & 2& 9& 2 & 10 & 10 &
			2 & \begin{tabular}[c]{@{}l@{}}2,39 times; 3,11 times \end{tabular} \\ 
			
			\emph{wine} & 3 & \begin{tabular}[c]{@{}l@{}}3\end{tabular} & 10&18 & 3 & 2 & 14 & 3& \begin{tabular}[c]{@{}l@{}}3,44 times; 2,6 times \end{tabular} \\ 
			\emph{seed} & 3 & 3& 10&12 &3 & 2 & 15& 3 &  \begin{tabular}[c]{@{}l@{}}3,50 times \end{tabular} \\ 
			
			\begin{tabular}[c]{@{}l@{}}\emph{\begin{tabular}[c]{@{}l@{}}\emph{bc}\end{tabular}} \end{tabular} & 2 & 9& 9 &15 & 3 &4&15& 3 & \begin{tabular}[c]{@{}l@{}}2,50 times\end{tabular} \\ 
			\emph{\begin{tabular}[c]{@{}l@{}}LS \end{tabular}} & 6 & 3 & 15&59 & 7 & 3 & 14 & 3 & \begin{tabular}[c]{@{}l@{}}2,50 times\end{tabular}\\ 
			\emph{magic} & 2 & 2 & 8&83 & 8 & 2 & 15& 2 & \begin{tabular}[c]{@{}l@{}}2,49 times; 5,1 time;\end{tabular} \\ \hline
		\end{tabular}
	}	
	
\end{table*}

\subsubsection{Discussion on clustering}
We have chosen six datasets for observing the performance of \emph{CNAK} in real-time. We found that four of them can be well clustered, namely, \emph{iris, breast-cancer, wine}, and \emph{seed}. We found an interesting observation in the \emph{wine} dataset. All data points are labeled to only one cluster after clustering by \emph{refining seed} method. In this situation, {\emph silhouette}  becomes invalid as it requires at least two clusters. This is confirmed by compactness measure (1.0) for this dataset. Notably, correct cluster number identification has a great impact on clustering accuracy. For example, We have found in our experiment that the \emph{iris} shows a high jump in \emph{ARI} from 0.57 to 0.89 when $k$ changes from 2 to 3. However, clustering results do not validate the outcome of the correct cluster number in \emph{magic} and \emph{land-satellite}. Table~\ref{tab:clustering_real1} suggests that \emph{CNAK} performs comparatively better than others.\\

\begin{table}[htb!]
	\centering
	\caption{Clustering results on real-world dataset with first two k appear in sorted score (ascending order) computed by \emph{CNAK}}
	\makebox[\linewidth]{
		\begin{tabular}{llccccccccccc}
			\hline
			\multirow{2}{*}{Dataset} & \multirow{2}{*}{Methods} & \multirow{2}{*}{K} & \multicolumn{2}{c}{ARI}  & \multicolumn{2}{c}{NMI} & \multicolumn{2}{c}{Homogeneity} & \multicolumn{2}{c}{Completeness} & \multicolumn{2}{c}{Silhouette} \\ \cline{4-13} 
			&  &  & Mode & SM & Mode & SM & Mode & SM & Mode & SM & Mode & SM \\ \hline
			\multirow{8}{*}{\emph{iris}} & \multirow{2}{*}{CNAK} & 3 & {\bf 0.88} &{\bf 0.89} &{\bf 0.86} &{\bf 0.86} &{\bf 0.86} &{\bf 0.86} &{\bf 0.86} & {\bf 0.86} &{\bf 0.75} & 0.63 \\  
			&  & 2 & 0.57 & 0.57 & 0.76 & 0.76 & 0.58 & 0.58 & 0.99& 0.99 & 0.70 & 0.62 \\  
			& \multirow{2}{*}{BA} & 3 & 0.83 & 0.84 & 0.83 & 0.84 & 0.83 & 0.83 & 0.84 & 0.84 & 0.74 & 0.68 \\  
			&  & 2 & 0.57 & 0.57 & 0.76 & 0.76 & 0.58 & 0.58 & 0.99 & 0.99 & 0.70 & 0.69 \\  
			& \multirow{2}{*}{RPKM} & 3 & 0.49 & 0.45 & 0.52 & 0.52 & 0.51 & 0.48& 0.52 & 0.56 & 0.33 & 0.35 \\  
			&  & 2 & 0.49 & 0.46 & 0.50 & 0.56 & 0.57 & 0.44 & 0.76 & 0.71 & 0.56 & 0.55 \\  
			& \multirow{2}{*}{RS} & 3 & 0.89 & 0.89 & 0.86 & 0.86 & 0.86 & 0.86 & 0.86& 0.86 & 0.75 & 0.75 \\ 
			&  & 2 & 0.57 & 0.57 & 0.76 & 0.76 & 0.58 & 0.58 & 0.99 & 0.99 & 0.70 & 0.70 \\ \hline
			\multirow{8}{*}{\emph{wine}} & \multirow{2}{*}{CNAK} & 3 & {\bf 0.80} &{\bf 0.78} & {\bf 0.80} &{\bf 0.78} &{\bf 0.80} &{\bf 0.79} &{\bf 0.80} &{\bf 0.78} &{\bf 0.49} &{\bf 0.44} \\ 
			&  & 2 & 0.37 & 0.37 & 0.50 & 0.50 & 0.40 & 0.39 & 0.64 & 0.64 & 0.41 & 0.39 \\ 
			& \multirow{2}{*}{BA} & 3 & 0.77 & 0.77 & 0.76 & 0.76 & 0.77 & 0.77 & 0.76 & 0.76 & 0.47 & 0.47 \\ 
			&  & 2 & 0.37 & 0.37 & 0.50 & 0.49 & 0.39 & 0.39 & 0.63 & 0.63 & 0.41 & 0.40 \\ 
			& \multirow{2}{*}{RPKM} & 3 & 0.33 & 0.32 & 0.40 & 0.37 & 0.37 & 0.34 & 044 & 0.46 & 0.34 & 0.23 \\ 
			&  & 2 & 0.34 & 0.26 & 0.45 & 0.33 & 0.35 & 0.25 & 0.56 & 0.54 & 0.36 & 0.32 \\ 
			& \multirow{2}{*}{RS} & 3 & 0.00 & 0.00 & 0.00 & 0.00 & 0.00 & 0.00 & 1.00 & 1.00 & - & - \\ 
			&  & 2 & 0.00 & 0.00 & 0.00 & 0.00 & 0.00 & 0.00 & 1.00 & 1.00 & - & - \\ \hline
			\multirow{8}{*}{\emph{seed}} & \multirow{2}{*}{\bf CNAK} & 3 &{\bf 0.71} & {\bf 0.69} &{\bf 0.67} &{\bf 0.66} &{\bf 0.67} &{\bf 0.66} &{\bf 0.67} &{\bf 0.66} & 0.45 & {\bf 0.52} \\ 
			&  & 2 & 0.44 & 0.44 & 0.53 & 0.53 & 0.42 & 0.42 & 0.67 & 0.67 & 0.29 & 0.24 \\ 
			& \multirow{2}{*}{BA} & 3 & 0.69 & 0.69 & 0.66 & 0.66 & 0.66 & 0.66 & 0.66 & 0.66 & {\bf 0.49} & 0.48 \\ 
			&  & 2 & 0.45 & 0.45 & 0.54 & 0.54 & 0.42 & 0.42 & 0.67 & 0.68 & 0.29 & 0.28 \\ 
			& \multirow{2}{*}{RPKM} & 3 & 0.48 & 0.49 & 0.45 & 0.54 & 0.51 & 0.50 & 0.58 & 0.59 & 0.39 & 0.22 \\ 
			&  & 2 & 0.48 & 0.44 & 0.56 & 0.53 & 0.44 & 0.41 & 0.72 & 0.69 & 0.21 & 0.23 \\ 
			& \multirow{2}{*}{RS} & 3 & 0.70 & 0.63 & 0.67 & 0.63 & 0.67 & 0.60 & 0.67 & 0.67 & 0.29 & 0.32 \\ 
			&  & 2 & 0.44 & 0.44 & 0.53 & 0.53 & 0.42 & 0.42 & 0.67 & 0.67 & 0.33 & 0.32 \\ \hline
			\multirow{8}{*}{\begin{tabular}[c]{@{}l@{}}\emph{breast-}\\ \emph{cancer}\end{tabular}} & \multirow{2}{*}{CNAK} & 2 &{\bf 0.85} & {\bf 0.85} &{\bf 0.75} &{\bf 0.75} &{\bf 0.74} &{\bf 0.74 }&{\bf 0.75} & {\bf 0.75} & 0.57 & 0.63 \\ 
			&  & 3 & 0.78 & 0.78 & 0.69 & 0.68 & 0.81 & 0.80 & 0.59 & 0.58 & 0.56 & 0.56 \\ 
			& \multirow{2}{*}{BA} & 2 &{\bf 0.85} &{\bf 0.85} &{\bf 0.75} &{\bf 0.75} &{\bf 0.74} &{\bf 0.74} &{\bf 0.75} &{\bf 0.75} &{\bf 0.69} &{\bf 0.69} \\ 
			&  & 3 & 0.81 & 0.81 & 0.73 & 0.73 & 0.87 & 0.87 & 0.62 & 0.62 & 0.58 & 0.58 \\ 
			& \multirow{2}{*}{RPKM} & 2 & 0.68 & 0.67 & 0.58 & 0.58 & 0.56 & 0.55 & 0.61 & 0.60 & 0.63 & 0.58 \\ 
			&  & 3 & 0.67 & 0.67 & 0.58 & 0.56 & 0.69 & 0.63 & 0.48 & 0.50 & 0.51 & 0.54 \\ 
			& \multirow{2}{*}{RS} & 2 & 0.85 & 0.85 & 0.75 & 0.75 & 0.74 & 0.74 & 0.75 & 0.75 & 0.69 & 0.69 \\ 
			&  & 3 & 0.79 & 0.79 & 0.70 & 0.70 & 0.82 & 0.82 & 0.59 & 0.59 & 0.58 & 0.58 \\ \hline
			
		\end{tabular}
	}
	\label{tab:clustering_real2}
\end{table}

\begin{table}[htb!]
	\centering
	\caption{clustering results on real-world Dataset for cluster number predicted by \emph{CNAK} and the ground truth.}
	\makebox[\linewidth]{
		\begin{tabular}{llccccccccccc}
			\hline
			\multirow{2}{*}{Dataset} & \multirow{2}{*}{Methods} & \multirow{2}{*}{K} & \multicolumn{2}{c}{ARI}  & \multicolumn{2}{c}{NMI} & \multicolumn{2}{c}{Homogeneity} & \multicolumn{2}{c}{Completeness} & \multicolumn{2}{c}{Silhouette} \\ \cline{4-13} 
			&  &  & Mode & SM & Mode & SM & Mode & SM & Mode & SM & Mode & SM \\ \hline
			\multirow{8}{*}{\begin{tabular}[c]{@{}l@{}}\emph{landsat}\\ \emph{satellite}\end{tabular}} & CNAK & 2 & 0.20 & 0.20 & 0.32 & 0.32 & 0.20 & 0.20 & 0.50 & 0.50 & 0.69 & 0.69 \\ 
			&  & 7 & 0.56 & 0.56 & 0.65 & 0.65 & 0.66 & 0.66 & 0.60 & 0.60 & 0.31 & 0.31 \\ 
			& \multirow{2}{*}{BA} & 2 & 0.19 & 0.19 & 0.30 & 0.30 & 0.19 & 0.19 & 0.47 & 0.47 & 0.37 & 0.37 \\ 
			&  & 7 & 0.49 & 0.49 & 0.55 & 0.55 & 0.58 & 0.58 & 0.53 & 0.53 & 0.29 & 0.29 \\ 
			& \multirow{2}{*}{RS} & 2 & 0.19 & 0.19 & 0.30 & 0.30 & 0.19 & 0.19 & 0.48 & 0.48 & 0.37 & 0.37 \\ 
			&  & 7 & 0.57 & 0.56 & 0.63 & 0.63 & 0.66 & 0.66 & 0.61 & 0.60 & 0.34 & 0.34 \\ 
			& \multirow{2}{*}{RPKM} & 2 & 0.19 & 0.14 & 0.30 & 0.29 & 0.18 & 0.16 & 0.60 & 0.54 & 0.33 & 0.37 \\ 
			&  & 7 & 0.33 & 0.30 & 0.45 & 0.40 & 0.43 & 0.37 & 0.47 & 0.44 & 0.17 & 0.17 \\ \hline
			\multirow{8}{*}{\emph{magic}} & \multirow{2}{*}{CNAK} & 4 & 0.11 & 0.11 & 0.12 & 0.12 & 0.17 & 0.17 & 0.09 & 0.09 & 0.17 & 0.17 \\ 
			&  & 2 & 0.10 & 0.10 & 0.06 & 0.06 & 0.06 & 0.06 & 0.05 & 0.05 & 0.22 & 0.21 \\ 
			& \multirow{2}{*}{BA} & 4 & 0.13 & 0.13 & 0.14 & 0.13 & 0.19 & 0.18 & 0.10 & 0.09 & 0.15 & 0.15 \\ 
			&  & 2 & 0.11 & 0.11 & 0.06 & 0.06 & 0.064 & 0.06 & 0.06 & 0.06 & 0.11 & 0.19 \\ 
			& \multirow{2}{*}{RS} & 4 & 0.12 & 0.09 & 0.10 & 0.08 & 0.12 & 0.07 & 0.08 & 0.09 & 0.17 & 0.19 \\ 
			&  & 2 & 0.11 & 0.08 & 0.11 & 0.09 & 0.07 & 0.05 & 0.17 & 0.19 & 0.30 & 0.27 \\ 
			& \multirow{2}{*}{RPKM} & 4 & 0.08 & 0.12 & 0.07 & 0.11 & 0.09 & 0.13 & 0.06 & 0.09 & 0.22 & 0.18 \\ 
			&  & 2 & 0.06 & 0.07 & 0.04 & 0.17 & 0.04 & 0.04 & 0.04 & 0.04 & 0.21 & 0.21 \\ \hline
			
		\end{tabular}
	}
	\label{tab:clustering_real1}
\end{table}

\subsubsection{Robustness against Additive white Gaussian noise}
We have examined the robustness of methods by incorporating additive Gaussian noise in 
\emph{breast-cancer} datasets. \emph{We use additive Gaussian noise with mean 0, and s is  a positive real number such that covariance matrix=$s \times I$ where $I$ represents the Identity matrix.} We set $s=3, 10$ and $30$ in our experiments. Table~\ref{tab:noisy_breast_cancer}~
shows that the performance of clustering methods decreases with increasing noise in the dataset. Also, our proposed method provides comparable quality of clustering with other methods in \emph{breast-cancer (3I and 10I)}. However, every metric except silhouette index for \emph{breast-cancer (30I)} indicates  that \emph{CNAK} is more tolerant than others in extreme noisy situation. 

	\begin{figure}[htb!]
	
	\centering
	
	\begin{multicols}{4}
		\subcaptionbox{breast-cancer\label{fig:BC-real}}{\includegraphics[width=0.25\textwidth]{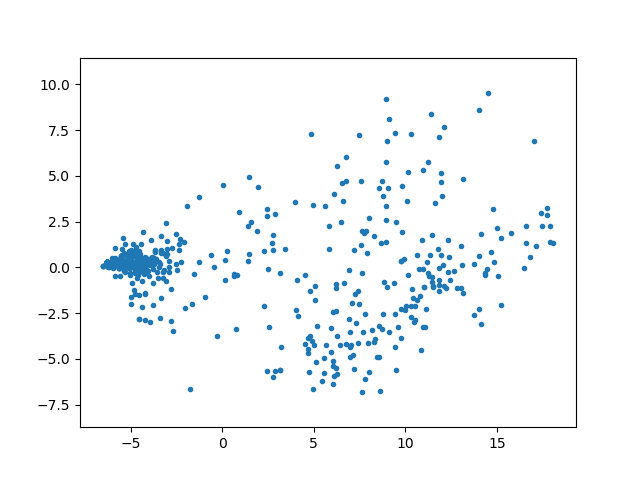}}
		\subcaptionbox{NC=3I\label{fig:BC-noise10}}{\includegraphics[width=0.25\textwidth]{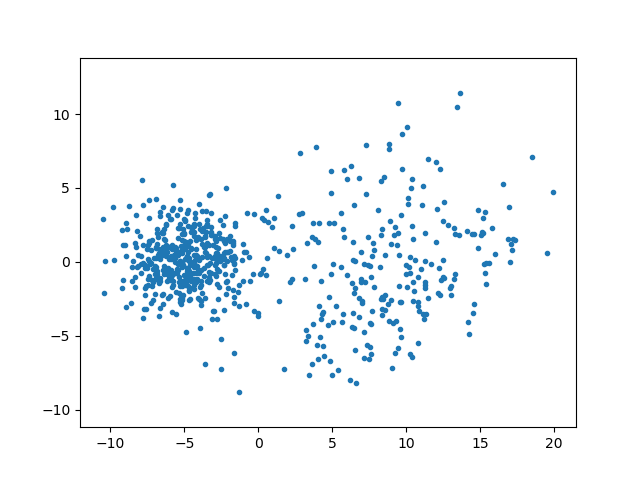}}
		\subcaptionbox{NC=10I\label{fig:BC-noise20}}{\includegraphics[width=0.25\textwidth]{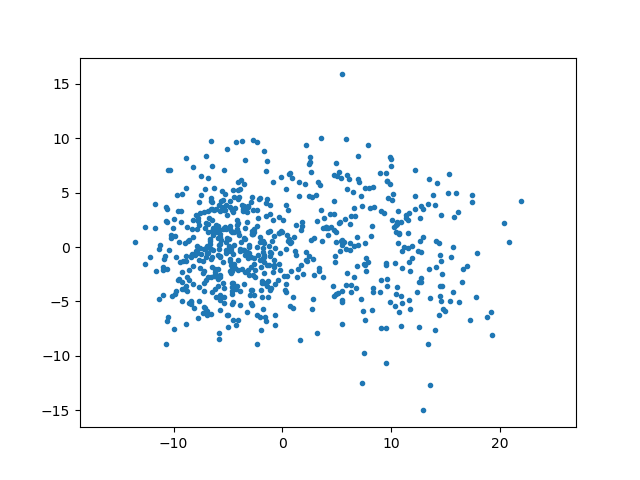}}
		\subcaptionbox{NC=30I\label{fig:BC-noise30}}{\includegraphics[width=0.25\textwidth]{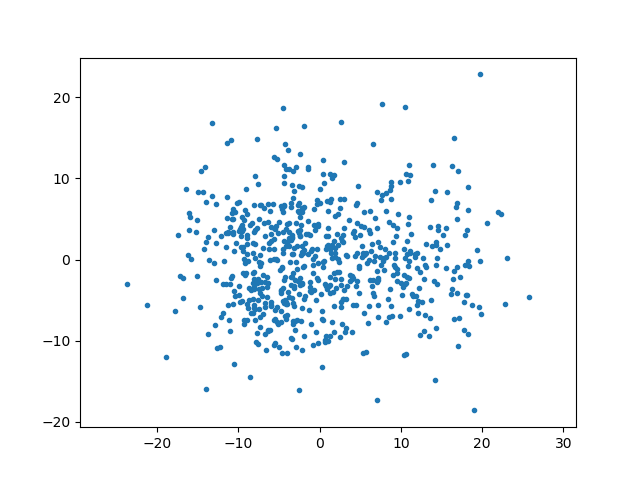}}
	\end{multicols}
	
	\caption{Scatter plot of the first two principal components of the breast-cancer dataset.  NC represents the covariance matrix of noise.}
	\label{fig:iris_bc}
\end{figure}

\begin{table}[htb!]
	\centering
	\caption{Prediction of cluster number for breast-cancer dataset with additive Gaussian noise}
	\makebox[\linewidth]{
		\begin{tabular}{lcccccccc}
			\hline
			\multirow{2}{*}{Dataset} & \multicolumn{8}{c}{Comparison of seven approached in cluster number prediction} \\ \cline{2-9} 
			& \multicolumn{1}{c}{CH} & \multicolumn{1}{c}{Jump} & \multicolumn{1}{c}{Hartigan} & \multicolumn{1}{c}{Curvature} & \multicolumn{1}{c}{Silhouette} & \multicolumn{1}{c}{Gap} & \multicolumn{1}{c}{$CV_{a}$} & \multicolumn{1}{c}{CNAK} \\ \hline
			{\begin{tabular}[c]{@{}l@{}}breast-cancer (3I)\end{tabular}} & 2 & 2 & 13 & 2 & 2 & 5 & 2 & 2 \\
			{\begin{tabular}[c]{@{}l@{}}breast-cancer (10I)\end{tabular}} & 2 & 2 & 15 & 2 & 2 & 2 & 2 & 2 \\
			{\begin{tabular}[c]{@{}l@{}}breast-cancer (30I)\end{tabular}} & 2 & 2 & 15 & 2 & 2 & 2 & 2 & 2 \\ \hline
		\end{tabular}
	}
\end{table}

\begin{table}[htb!]
	\caption{Clustering results on noisy breast-cancer dataset}
	\label{tab:noisy_breast_cancer}
	\makebox[\linewidth]{
		\begin{tabular}{llllllllllll}
			\hline
			\multirow{2}{*}{Dataset} & \multirow{2}{*}{Methods} & \multicolumn{2}{c}{ARI} & \multicolumn{2}{c}{NMI} & \multicolumn{2}{c}{Homogeneity} & \multicolumn{2}{c}{Completeness} & \multicolumn{2}{c}{Silhouette} \\
			&  & \multicolumn{1}{c}{Mode} & \multicolumn{1}{c}{SM} & \multicolumn{1}{c}{Mode} & \multicolumn{1}{c}{SM} & \multicolumn{1}{c}{Mode} & \multicolumn{1}{c}{SM} & \multicolumn{1}{c}{Mode} & \multicolumn{1}{c}{SM} & Mode & SM \\ \hline
			\multirow{4}{*}{\begin{tabular}[c]{@{}l@{}}breast\\-cancer (3I)\end{tabular}} & CNAK & {\bf 0.84} & {\bf 0.84} & {\bf 0.74} & {\bf 0.74} & {\bf 0.73} & {\bf 0.73} & {\bf 0.75} & {\bf 0.75} & 0.42 & 0.42 \\
			& BA & 0.83 & 0.83 & 0.73 & 0.73 & 0.73 & 0.73 & 0.74 & 0.74 & {\bf 0.62} & {\bf 0.62} \\
			& RS & 0.83 & 0.83 & 0.73 & 0.73 & 0.73 & 0.73 & 0.74 & 0.74 & {\bf 0.62} & {\bf0.62} \\
			& RPKM & 0.71 & 0.66 & 0.62 &0.56  & 0.61 & 0.54 & 0.60 & 0.58 & 0.49 &0.47  \\ \hline
			\multirow{4}{*}{\begin{tabular}[c]{@{}l@{}}breast\\-cancer (10I)\end{tabular}} & CNAK & {\bf0.79} & {\bf 0.78} & {\bf 0.67} & {\bf 0.67} & {\bf 0.67} & {\bf 0.67} & {\bf 0.67} & {\bf 0.68} & 0.27 & 0.27 \\
			& BA & 0.76 & 0.75 & 0.63 & 0.63 & 0.63 & 0.63 & 0.64 & 0.63 & {\bf 0.52} & {\bf0.51} \\
			& RS & 0.75 & 0.75 & 0.63 & 0.63 & 0.63 & 0.63 & 0.63 & 0.63 & 0.51 & {\bf 0.51} \\
			& RPKM & 0.72 &  0.67 & 0.62 & 0.56 & 0.55 & 0.54 &0.61  &0.56 & 0.40 &0.39  \\ \hline
			\multirow{4}{*}{\begin{tabular}[c]{@{}l@{}}breast\\-cancer (30I)\end{tabular}} & CNAK & 
			{\bf 0.56} & {\bf 0.56} & {\bf 0.43} & {\bf 0.44} & {\bf 0.44} & {\bf 0.44} & {\bf 0.43} & {\bf 0.44} & { 0.16} &{ 0.16} \\
			& BA & 0.49 & 0.48 & 0.37 & 0.37 & 0.38 & 0.38 & 0.37 & 0.37 & {\bf 0.32} & 0.31 \\
			& RS & 0.49 & 0.49 & 0.37 & 0.37 & 0.38 & 0.38 & 0.37 & 0.37 & {\bf 0.32} & {\bf0.32} \\
			& RPKM & 0.36 & 0.27 & 0.25 &0.18  & 0.25 &0.18  & 0.25 & 0.18 & 0.12 & 0.11 \\ \hline
		\end{tabular}
	}
\end{table}

\section{Conclusion}\label{sec:conclusion}
In this paper, we propose a novel cluster number assisted \emph{k-means} (\emph{CNAK}) algorithm. The proposed method consists of two steps. Initially, the number of clusters is estimated. In the second step, the clustering of data points is obtained. Automation in the determination of the cluster number determination is a major improvement in our work. In this paper, we investigate the applicability of our proposed work to handle a large scale dataset in an increasing number of instances as well as increasing dimensions. We study a few pertinent issues related to clustering. We design many simulations to discuss those issues. 
Our investigation reveals that the \emph{CNAK} can predict a single cluster as well as multiple clusters. We find \emph{CNAK} is robust against cluster imbalance as suggested by the results of clustering on \emph{sim-4}.  Moreover, it is efficient in handling a large scale dataset. But, its performance declines when the ``between-cluster separation" is small.	
Also, it is unable to detect  any shape clusters. Though it can indicate the existing hierarchical structure via a detailed analysis of the plot of ``k vs. score", the performance of clustering decreases with the increase in the depth of the hierarchy.
The proposed algorithm \emph{CNAK} is a \emph{Monte-Carlo} simulation based method. Hence, determination of the size of a sampled dataset is a very crucial step and a key factor for the successful performance of the algorithm. However, with small scale data, we may work with a sample size of a higher fraction of   dataset. But, this is not applicable, when the size of dataset becomes very large. 
We have proposed a heuristic to determine the size of a sampled dataset for a given dataset whose prediction for large scale dataset is feasible. The drop in computational time for the large dataset is governed by the small size of the sampled dataset in our proposed method. However, we observed that our heuristic fails when the dataset contains many clusters. In the future, we intend to apply our algorithm on a real-world, large-scale dataset for data mining.

\section{Acknowledgments}
This research did not receive any specific grant from funding agencies in the public, commercial, or not-for-profit sectors.

	\section{References}
	\bibliographystyle{model5-names}
	\bibliography{CNAK_arxiv.bib}

\begin{thebibliography}{39}
\expandafter\ifx\csname natexlab\endcsname\relax\def\natexlab#1{#1}\fi
\providecommand{\bibinfo}[2]{#2}
\ifx\xfnm\relax \def\xfnm[#1]{\unskip,\space#1}\fi
\bibitem[{Arthur \& Vassilvitskii(2007)}]{Arthur:2007}
\bibinfo{author}{Arthur, D.}, \& \bibinfo{author}{Vassilvitskii, S.}
  (\bibinfo{year}{2007}).
\newblock \bibinfo{title}{{K-means++: The Advantages of Careful Seeding}}.
\newblock In {\it \bibinfo{booktitle}{Symposium on Discrete Algorithms}\/} (pp.
  \bibinfo{pages}{1027--1035}).
\newblock \bibinfo{publisher}{SIAM}.
\bibitem[{Bradley \& Fayyad(1998)}]{Bradley1998RefiningIP}
\bibinfo{author}{Bradley, P.~S.}, \& \bibinfo{author}{Fayyad, U.~M.}
  (\bibinfo{year}{1998}).
\newblock \bibinfo{title}{{Refining Initial Points for K-Means Clustering}}.
\newblock In {\it \bibinfo{booktitle}{International Conference on Machine
  Learning}\/} (pp. \bibinfo{pages}{91--99}).
\newblock \bibinfo{publisher}{Morgan Kaufmann Publishers Inc.}
\bibitem[{Calinski \& Harabasz(1974)}]{CH}
\bibinfo{author}{Calinski, T.}, \& \bibinfo{author}{Harabasz, J.}
  (\bibinfo{year}{1974}).
\newblock \bibinfo{title}{{A dendrite method for cluster analysis}}.
\newblock {\it \bibinfo{journal}{Communications in Statistics}\/},  {\it
  \bibinfo{volume}{3}\/}, \bibinfo{pages}{1--27}.
\bibitem[{Capo et~al.(2017)Capo, Pérez \& Lozano}]{CAPO2017}
\bibinfo{author}{Capo, M.}, \bibinfo{author}{Pérez, A.}, \&
  \bibinfo{author}{Lozano, J.~A.} (\bibinfo{year}{2017}).
\newblock \bibinfo{title}{{An efficient approximation to the K-means clustering
  for massive data}}.
\newblock {\it \bibinfo{journal}{Knowledge-Based Systems}\/},  {\it
  \bibinfo{volume}{117}\/}, \bibinfo{pages}{56--69}.
\bibitem[{Chang \& Yeung(2008)}]{pathbased_spiral}
\bibinfo{author}{Chang, H.}, \& \bibinfo{author}{Yeung, D.-Y.}
  (\bibinfo{year}{2008}).
\newblock \bibinfo{title}{Robust path-based spectral clustering}.
\newblock {\it \bibinfo{journal}{Pattern Recognition}\/},  {\it
  \bibinfo{volume}{41}\/}, \bibinfo{pages}{191 -- 203}.
\bibitem[{Chiang \& Mirkin(2010)}]{R3_Chiang2010}
\bibinfo{author}{Chiang, M. M.-T.}, \& \bibinfo{author}{Mirkin, B.}
  (\bibinfo{year}{2010}).
\newblock \bibinfo{title}{Intelligent choice of the number of clusters in
  k-means clustering: An experimental study with different cluster spreads}.
\newblock {\it \bibinfo{journal}{Journal of Classification}\/},  {\it
  \bibinfo{volume}{27}\/}, \bibinfo{pages}{3--40}.
\bibitem[{Davidson \& Satyanarayana(2003)}]{Davidson2003}
\bibinfo{author}{Davidson, I.}, \& \bibinfo{author}{Satyanarayana, A.}
  (\bibinfo{year}{2003}).
\newblock \bibinfo{title}{{Speeding up k-means Clustering by Bootstrap
  Averaging}}.
\newblock In {\it \bibinfo{booktitle}{ICDM Workshop on Clustering Large Data
  Sets}\/} (pp. \bibinfo{pages}{16--25}).
\newblock \bibinfo{publisher}{IEEE}.
\bibitem[{Davies \& Bouldin(1979)}]{DB}
\bibinfo{author}{Davies, D.~L.}, \& \bibinfo{author}{Bouldin, D.~W.}
  (\bibinfo{year}{1979}).
\newblock \bibinfo{title}{{A Cluster Separation Measure}}.
\newblock {\it \bibinfo{journal}{Transactions on Pattern Analysis and Machine
  Intelligence}\/},  {\it \bibinfo{volume}{1}\/}, \bibinfo{pages}{224--227}.
\bibitem[{Dheeru \& Karra~Taniskidou(2017)}]{Dua:2017}
\bibinfo{author}{Dheeru, D.}, \& \bibinfo{author}{Karra~Taniskidou, E.}
  (\bibinfo{year}{2017}).
\newblock \bibinfo{title}{{UCI Machine Learning Repository}}.
\bibitem[{Dudoit \& Fridlyand(2002)}]{Dudoit_02}
\bibinfo{author}{Dudoit, S.}, \& \bibinfo{author}{Fridlyand, J.}
  (\bibinfo{year}{2002}).
\newblock \bibinfo{title}{A prediction-based resampling method for estimating
  the number of clusters in a dataset.}
\newblock {\it \bibinfo{journal}{Genome biology}\/},  {\it
  \bibinfo{volume}{3}\/}.
\bibitem[{Edmonds \& Karp(1972)}]{kuhn-munkre}
\bibinfo{author}{Edmonds, J.}, \& \bibinfo{author}{Karp, R.~M.}
  (\bibinfo{year}{1972}).
\newblock \bibinfo{title}{Theoretical improvements in algorithmic efficiency
  for network flow problems}.
\newblock {\it \bibinfo{journal}{J. ACM}\/},  {\it \bibinfo{volume}{19}\/},
  \bibinfo{pages}{248--264}.
\bibitem[{Ester et~al.(1996)Ester, Kriegel, Sander \& Xu}]{DBSCAN}
\bibinfo{author}{Ester, M.}, \bibinfo{author}{Kriegel, H.-P.},
  \bibinfo{author}{Sander, J.}, \& \bibinfo{author}{Xu, X.}
  (\bibinfo{year}{1996}).
\newblock \bibinfo{title}{{A Density-based Algorithm for Discovering Clusters a
  Density-based Algorithm for Discovering Clusters in Large Spatial Databases
  with Noise}}.
\newblock In {\it \bibinfo{booktitle}{International Conference on Knowledge
  Discovery and Data Mining}\/} (pp. \bibinfo{pages}{226--231}).
\newblock \bibinfo{publisher}{AAAI Press}.
\bibitem[{Estiri et~al.(2018)Estiri, Omran \& Murphy}]{kluster2018}
\bibinfo{author}{Estiri, H.}, \bibinfo{author}{Omran, B.~A.}, \&
  \bibinfo{author}{Murphy, S.~N.} (\bibinfo{year}{2018}).
\newblock \bibinfo{title}{kluster: An efficient scalable procedure for
  approximating the number of clusters in unsupervised learning}.
\newblock {\it \bibinfo{journal}{Big Data Research}\/},  {\it
  \bibinfo{volume}{13}\/}, \bibinfo{pages}{38 -- 51}.
\bibitem[{Fang \& Wang(2012)}]{instability2012}
\bibinfo{author}{Fang, Y.}, \& \bibinfo{author}{Wang, J.}
  (\bibinfo{year}{2012}).
\newblock \bibinfo{title}{Selection of the number of clusters via the bootstrap
  method}.
\newblock {\it \bibinfo{journal}{Computational Statistics \& Data Analysis}\/},
   {\it \bibinfo{volume}{56}\/}, \bibinfo{pages}{468 -- 477}.
\bibitem[{Fr\"anti et~al.(2006)Fr\"anti, Virmajoki \& Hautam\"aki}]{DIMsets}
\bibinfo{author}{Fr\"anti, P.}, \bibinfo{author}{Virmajoki, O.}, \&
  \bibinfo{author}{Hautam\"aki, V.} (\bibinfo{year}{2006}).
\newblock \bibinfo{title}{Fast agglomerative clustering using a k-nearest
  neighbor graph}.
\newblock {\it \bibinfo{journal}{IEEE Trans. on Pattern Analysis and Machine
  Intelligence}\/},  {\it \bibinfo{volume}{28}\/}, \bibinfo{pages}{1875--1881}.
\bibitem[{Fu \& Medico(2007)}]{Flame}
\bibinfo{author}{Fu, L.}, \& \bibinfo{author}{Medico, E.}
  (\bibinfo{year}{2007}).
\newblock \bibinfo{title}{Flame, a novel fuzzy clustering method for the
  analysis of dna microarray data}.
\newblock {\it \bibinfo{journal}{BMC Bioinformatics}\/},  {\it
  \bibinfo{volume}{8}\/}, \bibinfo{pages}{3}.
\bibitem[{Gionis et~al.(2007)Gionis, Mannila \& Tsaparas}]{Aggregation}
\bibinfo{author}{Gionis, A.}, \bibinfo{author}{Mannila, H.}, \&
  \bibinfo{author}{Tsaparas, P.} (\bibinfo{year}{2007}).
\newblock \bibinfo{title}{Clustering aggregation}.
\newblock {\it \bibinfo{journal}{ACM Trans. Knowl. Discov. Data}\/},  {\it
  \bibinfo{volume}{1}\/}.
\bibitem[{Gupta et~al.(2018)Gupta, Datta \& Das}]{ISI_LL_LML2018}
\bibinfo{author}{Gupta, A.}, \bibinfo{author}{Datta, S.}, \&
  \bibinfo{author}{Das, S.} (\bibinfo{year}{2018}).
\newblock \bibinfo{title}{Fast automatic estimation of the number of clusters
  from the minimum inter-center distance for k-means clustering}.
\newblock {\it \bibinfo{journal}{Pattern Recognition Letters}\/},  {\it
  \bibinfo{volume}{116}\/}, \bibinfo{pages}{72 -- 79}.
\bibitem[{Hancer \& Karaboga(2017)}]{survey_cluster_number2017}
\bibinfo{author}{Hancer, E.}, \& \bibinfo{author}{Karaboga, D.}
  (\bibinfo{year}{2017}).
\newblock \bibinfo{title}{A comprehensive survey of traditional, merge-split
  and evolutionary approaches proposed for determination of cluster number}.
\newblock {\it \bibinfo{journal}{Swarm and Evolutionary Computation}\/},  {\it
  \bibinfo{volume}{32}\/}, \bibinfo{pages}{49 -- 67}.
\bibitem[{Hartigan(1975)}]{hartigan}
\bibinfo{author}{Hartigan, J.~A.} (\bibinfo{year}{1975}).
\newblock {\it \bibinfo{title}{{Clustering Algorithms}}\/}.
\newblock (\bibinfo{edition}{99th} ed.).
\newblock \bibinfo{publisher}{John Wiley \& Sons, Inc.}
\bibitem[{Hinneburg \& Keim(1998)}]{DENCLUE}
\bibinfo{author}{Hinneburg, A.}, \& \bibinfo{author}{Keim, D.~A.}
  (\bibinfo{year}{1998}).
\newblock \bibinfo{title}{{An Efficient Approach to Clustering in Large
  Multimedia Databases with Noise}}.
\newblock In {\it \bibinfo{booktitle}{International Conference on Knowledge
  Discovery and Data Mining}\/} (pp. \bibinfo{pages}{58--65}).
\newblock \bibinfo{publisher}{AAAI Press}.
\bibitem[{Horibe(1985)}]{NMI}
\bibinfo{author}{Horibe, Y.} (\bibinfo{year}{1985}).
\newblock \bibinfo{title}{Entropy and correlation}.
\newblock {\it \bibinfo{journal}{IEEE Transactions on Systems, Man, and
  Cybernetics}\/},  {\it \bibinfo{volume}{SMC-15}\/},
  \bibinfo{pages}{641--642}.
\bibitem[{Hubert \& Arabie(1985)}]{Hubert1985}
\bibinfo{author}{Hubert, L.}, \& \bibinfo{author}{Arabie, P.}
  (\bibinfo{year}{1985}).
\newblock \bibinfo{title}{{Comparing partitions}}.
\newblock {\it \bibinfo{journal}{Journal of Classification}\/},  {\it
  \bibinfo{volume}{2}\/}, \bibinfo{pages}{193--218}.
\bibitem[{Jain \& Law(2005)}]{Jain}
\bibinfo{author}{Jain, A.~K.}, \& \bibinfo{author}{Law, M. H.~C.}
  (\bibinfo{year}{2005}).
\newblock \bibinfo{title}{Data clustering: A user's dilemma}.
\newblock In \bibinfo{editor}{S.~K. Pal}, \bibinfo{editor}{S.~Bandyopadhyay},
  \& \bibinfo{editor}{S.~Biswas} (Eds.), {\it \bibinfo{booktitle}{Pattern
  Recognition and Machine Intelligence}\/} (pp. \bibinfo{pages}{1--10}).
\bibitem[{Jr.(1963)}]{Hierarchy1963}
\bibinfo{author}{Jr., J. H.~W.} (\bibinfo{year}{1963}).
\newblock \bibinfo{title}{Hierarchical grouping to optimize an objective
  function}.
\newblock {\it \bibinfo{journal}{Journal of the American Statistical
  Association}\/},  {\it \bibinfo{volume}{58}\/}, \bibinfo{pages}{236--244}.
\bibitem[{Lloyd(1982)}]{Lloyd57}
\bibinfo{author}{Lloyd, S.} (\bibinfo{year}{1982}).
\newblock \bibinfo{title}{{Least squares quantization in PCM}}.
\newblock {\it \bibinfo{journal}{Transactions on Information Theory}\/},  {\it
  \bibinfo{volume}{28}\/}, \bibinfo{pages}{129--137}.
\bibitem[{Ng et~al.(2001)Ng, Jordan \& Weiss}]{onspectral2001}
\bibinfo{author}{Ng, A.~Y.}, \bibinfo{author}{Jordan, M.~I.}, \&
  \bibinfo{author}{Weiss, Y.} (\bibinfo{year}{2001}).
\newblock \bibinfo{title}{On spectral clustering: Analysis and an algorithm}.
\newblock In {\it \bibinfo{booktitle}{ADVANCES IN NEURAL INFORMATION PROCESSING
  SYSTEMS}\/} (pp. \bibinfo{pages}{849--856}).
\newblock \bibinfo{publisher}{MIT Press}.
\bibitem[{Rezaei \& Fr\"anti(2016)}]{UnbalanceSet}
\bibinfo{author}{Rezaei, M.}, \& \bibinfo{author}{Fr\"anti, P.}
  (\bibinfo{year}{2016}).
\newblock \bibinfo{title}{Set-matching methods for external cluster validity}.
\newblock {\it \bibinfo{journal}{IEEE Trans. on Knowledge and Data
  Engineering}\/},  {\it \bibinfo{volume}{28}\/}, \bibinfo{pages}{2173--2186}.
\bibitem[{Rosenberg \& Hirschberg(2007)}]{vMeasure}
\bibinfo{author}{Rosenberg, A.}, \& \bibinfo{author}{Hirschberg, J.}
  (\bibinfo{year}{2007}).
\newblock \bibinfo{title}{{V-Measure: A Conditional Entropy-Based External
  Cluster Evaluation Measure}}.
\newblock In {\it \bibinfo{booktitle}{Joint Conference on Empirical Methods in
  Natural Language Processing and Computational Natural Language Learning}\/}
  (pp. \bibinfo{pages}{410--420}).
\bibitem[{Rousseeuw(1987)}]{sil}
\bibinfo{author}{Rousseeuw, P.~J.} (\bibinfo{year}{1987}).
\newblock \bibinfo{title}{{Silhouettes: A graphical aid to the interpretation
  and validation of cluster analysis}}.
\newblock {\it \bibinfo{journal}{Journal of Computational and Applied
  Mathematics}\/},  {\it \bibinfo{volume}{20}\/}, \bibinfo{pages}{53--65}.
\bibitem[{Sugar \& James(2003)}]{jump}
\bibinfo{author}{Sugar, C.~A.}, \& \bibinfo{author}{James, G.~M.}
  (\bibinfo{year}{2003}).
\newblock \bibinfo{title}{{Finding the Number of Clusters in a Dataset: An
  Information-Theoretic Approach}}.
\newblock {\it \bibinfo{journal}{Journal of the American Statistical
  Association}\/},  {\it \bibinfo{volume}{98}\/}, \bibinfo{pages}{750--763}.
\bibitem[{Tibshirani et~al.(2001)Tibshirani, Walther \& Hastie}]{gap}
\bibinfo{author}{Tibshirani, R.}, \bibinfo{author}{Walther, G.}, \&
  \bibinfo{author}{Hastie, T.} (\bibinfo{year}{2001}).
\newblock \bibinfo{title}{{Estimating the number of clusters in a data set via
  the gap statistic}}.
\newblock {\it \bibinfo{journal}{Journal of the Royal Statistical Society:
  Series B (Statistical Methodology)}\/},  {\it \bibinfo{volume}{63}\/},
  \bibinfo{pages}{411--423}.
\bibitem[{Tryfos(2009)}]{sampling}
\bibinfo{author}{Tryfos, P.} (\bibinfo{year}{2009}).
\newblock {\it \bibinfo{title}{{Simple random samples and their
  properties}}\/}.
\newblock \bibinfo{type}{Technical Report} University of York.
\bibitem[{Veenman et~al.(2002)Veenman, Reinders \& Backer}]{D31_R15}
\bibinfo{author}{Veenman, C.~J.}, \bibinfo{author}{Reinders, M. J.~T.}, \&
  \bibinfo{author}{Backer, E.} (\bibinfo{year}{2002}).
\newblock \bibinfo{title}{A maximum variance cluster algorithm}.
\newblock {\it \bibinfo{journal}{IEEE Transactions on Pattern Analysis and
  Machine Intelligence}\/},  {\it \bibinfo{volume}{24}\/},
  \bibinfo{pages}{1273--1280}.
\bibitem[{Wang(2010)}]{CV_A}
\bibinfo{author}{Wang, J.} (\bibinfo{year}{2010}).
\newblock \bibinfo{title}{{Consistent selection of the number of clusters via
  crossvalidation}}.
\newblock {\it \bibinfo{journal}{Biometrika}\/},  {\it \bibinfo{volume}{97}\/},
  \bibinfo{pages}{893--904}.
\bibitem[{Xu et~al.(1998)Xu, Ester, Kriegel \& Sander}]{DBCLASD}
\bibinfo{author}{Xu, X.}, \bibinfo{author}{Ester, M.},
  \bibinfo{author}{Kriegel, H.~P.}, \& \bibinfo{author}{Sander, J.}
  (\bibinfo{year}{1998}).
\newblock \bibinfo{title}{{A distribution-based clustering algorithm for mining
  in large spatial databases}}.
\newblock In {\it \bibinfo{booktitle}{International Conference on Data
  Engineering}\/} (pp. \bibinfo{pages}{324--331}).
\newblock \bibinfo{publisher}{IEEE}.
\bibitem[{Zahn(1971)}]{Compound}
\bibinfo{author}{Zahn, C.~T.} (\bibinfo{year}{1971}).
\newblock \bibinfo{title}{Graph-theoretical methods for detecting and
  describing gestalt clusters}.
\newblock {\it \bibinfo{journal}{IEEE Transactions on Computers}\/},  {\it
  \bibinfo{volume}{C-20}\/}, \bibinfo{pages}{68--86}.
\bibitem[{Zhang et~al.(2017)Zhang, Ma\'ndziuk, Quek \& Goh}]{Curvature}
\bibinfo{author}{Zhang, Y.}, \bibinfo{author}{Ma\'ndziuk, J.},
  \bibinfo{author}{Quek, C.~H.}, \& \bibinfo{author}{Goh, B.~W.}
  (\bibinfo{year}{2017}).
\newblock \bibinfo{title}{{Curvature-based method for determining the number of
  clusters}}.
\newblock {\it \bibinfo{journal}{Information Sciences}\/},  {\it
  \bibinfo{volume}{415-416}\/}, \bibinfo{pages}{414--428}.
\bibitem[{del Águila \& González-Ramírez(2014)}]{SS}
\bibinfo{author}{del Águila, M.~R.}, \& \bibinfo{author}{González-Ramírez,
  A.} (\bibinfo{year}{2014}).
\newblock \bibinfo{title}{Sample size calculation}.
\newblock {\it \bibinfo{journal}{Allergologia et Immunopathologia}\/},  {\it
  \bibinfo{volume}{42}\/}, \bibinfo{pages}{485 -- 492}.

\end{thebibliography}

\end{document}